%% file: main.tex
\documentclass[10pt,twocolumn,letterpaper]{article}

\usepackage[pagenumbers]{cvpr} %

\input{mymacros.tex}

\usepackage[accsupp]{axessibility}  %

\newcommand{\ourDatasetShortName}{MONDAY\xspace}

\newcommand{\ourDatasetFullBoldfaceName}{\textbf{M}obile \textbf{O}S \textbf{N}avigation Task \textbf{D}ataset for \textbf{A}gents from \textbf{Y}ouTube\xspace}

\newcommand{\ourTitle}{Scalable Video-to-Dataset Generation for Cross-Platform Mobile Agents}

\definecolor{customgreen}{HTML}{17a94c}
\definecolor{customred}{HTML}{fa2d2d}
\newcommand{\colorcmark}{{\color{customgreen}\ding{51}}}
\newcommand{\colorxmark}{{\color{customred}\ding{55}}}

\definecolor{cvprblue}{rgb}{0.21,0.49,0.74}
\usepackage[pagebackref,breaklinks,colorlinks,allcolors=cvprblue]{hyperref}

\def\paperID{1515} %
\def\confName{CVPR}
\def\confYear{2025}

\title{\ourTitle}

\newlength{\auSpacing}
\author{
Yunseok Jang$^{*\dag}$ \hspace{\auSpacing}
Yeda Song$^{*\dag}$ \hspace{\auSpacing}
Sungryull Sohn$^{\ddag}$ \hspace{\auSpacing}
Lajanugen Logeswaran$^{\ddag}$
\\
Tiange Luo$^{\dag}$ \hspace{\auSpacing}
Dong-Ki Kim$^{\ddag}$ \hspace{\auSpacing}
Kyunghoon Bae$^{\ddag}$ \hspace{\auSpacing}
Honglak Lee$^{\dag\ddag}$
\\
{
    $^{\dag}$University of Michigan \quad\quad\quad
    $^{\ddag}$LG AI Research
}
\\
\small{\texttt{$^{\dag} $\{\href{mailto:yunseokj@umich.edu}{yunseokj}, \href{mailto:yedasong@umich.edu}{yedasong}, \href{mailto:tiangel@umich.edu}{tiangel}, \href{mailto:honglak@umich.edu}{honglak}\}@umich.edu}}
\\
\small{\texttt{$^{\ddag} $\{\href{mailto:srsohn@lgresearch.ai}{srsohn}, \href{mailto:llajan@lgresearch.ai}{llajan},
\href{mailto:dkkim@lgresearch.ai}{dkkim},
\href{mailto:k.bae@lgresearch.ai}{k.bae}, \href{mailto:honglak@lgresearch.ai}{honglak}\}@lgresearch.ai}}
\\
\href{https://monday-dataset.github.io}{https://monday-dataset.github.io}
}

\begin{document}
\maketitle

\def\thefootnote{*}\footnotetext{Equal contribution}\def\thefootnote{\arabic{footnote}}

\begin{abstract}
\input{review/text/abstract.tex}
\end{abstract}

\section{Introduction}
\label{mobvid:sec:introduction}
\input{review/text/intro.tex}

\section{Related Work}
\label{mobvid:sec:related}
\input{review/text/related.tex}

\section{\ourDatasetShortName}
\label{mobvid:sec:dataset}

\input{review/text/dataset.tex}

\section{Experiments}
\label{mobvid:sec:experiments}
\input{review/text/experiments.tex}

\section{Conclusion}
\label{mobvid:sec:conclusion}
\input{review/text/conclusion.tex}

\newpage
\section*{Acknowledgements}
\label{mobvid:sec:acknowledgements}
We thank Jaekyeom Kim, Jae-Won Chung and Paula Suh for constructive feedbacks.
This work was supported in part by LG AI Research, OpenAI Researcher Access Program, and Kwanjeong Educational Foundation Scholarship.

{
    \small
    \bibliographystyle{ieeenat_fullname}
    \bibliography{mobile_os}
}

\clearpage
\appendix

\setcounter{page}{1}
\twocolumn[\textbf{\Large Supplementary Material of ``\ourTitle''} \vspace{2em}]

\renewcommand{\thetable}{\Alph{table}}
\renewcommand{\thefigure}{\Alph{figure}}
\renewcommand{\thesection}{\Alph{section}}
\renewcommand{\theequation}{\Alph{equation}}
\setcounter{figure}{0}
\setcounter{table}{0}
\setcounter{equation}{0}
\setcounter{footnote}{0}

\section{More Statistics about \ourDatasetShortName Dataset}
\label{mobvid:suppsec:additional_dataset_details}
\input{review/supp/additional_dataset_details.tex}

\section{Details about Video Collection}
\label{mobvid:suppsec:video_collection}
\input{review/supp/video_collection.tex}

\section{Details about \ourDatasetShortName Framework}
\label{mobvid:suppsec:pipeline_details}
\input{review/supp/pipeline_details.tex}

\section{Annotation of the Evaluation Dataset}
\label{mobvid:suppsec:evaluation_dataset_collection}
\input{review/supp/evaluation_dataset_collection.tex}

\section{More Examples from Dataset Collection Method Evaluation}
\label{mobvid:suppsec:additional_qualitative_examples}
\input{review/supp/additional_qualitative_examples.tex}

\section{Human Evaluation of the \ourDatasetShortName Dataset}
\label{mobvid:suppsec:human_evaluation}
\input{review/supp/human_evaluation.tex}

\newpage
\section{Details about Model Training Experiment}
\label{mobvid:suppsec:model_finetuning}

\input{review/supp/model_finetuning.tex}

\section{Expanded Related Work}
\label{mobvid:suppsec:expanded_related}
\input{review/supp/expanded_related.tex}

\section{Episode Examples}
\label{mobvid:suppsec:episode_examples}

\input{review/supp/episode_examples.tex}

\end{document}

%% file: mymacros.tex
\usepackage{color}
\usepackage{amsmath}
\usepackage{amssymb}
\usepackage{amsfonts}
\usepackage{multirow, varwidth, rotating}
\usepackage{epsfig}
\usepackage{verbatim}
\usepackage{dsfont}
\makeatletter
\newif\if@restonecol
\makeatother
\usepackage{xr}
\usepackage{booktabs}
\usepackage{comment}
\usepackage{paralist}
\usepackage{listings}
\usepackage{xcolor}
\usepackage{xspace}

\usepackage{subcaption}
\usepackage{pifont}%

\DeclareRobustCommand\onedot{\futurelet\@let@token\@onedot}
\def\onedot{.}
\def\eg{\emph{e.g}\onedot}

\def\etal{\emph{et al}\onedot}

%% file: review/text/abstract.tex
Recent advancements in Large Language Models (LLMs) and Vision-Language Models (VLMs) have sparked significant interest in developing GUI visual agents. 
We introduce \ourDatasetShortName (\ourDatasetFullBoldfaceName), a large-scale dataset of 313K annotated frames from 20K instructional videos capturing diverse real-world mobile OS navigation across multiple platforms. 
Models that include \ourDatasetShortName in their pre-training phases demonstrate robust cross-platform generalization capabilities, consistently outperforming models trained on existing single OS datasets while achieving an average performance gain of 18.11\%p on an unseen mobile OS platform.
To enable continuous dataset expansion as mobile platforms evolve, we present an automated framework that leverages publicly available video content to create comprehensive task datasets without manual annotation. 
Our framework comprises robust OCR-based scene detection (95.04\% F1-score), near-perfect UI element detection (99.87\% hit ratio), and novel multi-step action identification to extract reliable action sequences across diverse interface configurations. 
We contribute both the \ourDatasetShortName dataset and our automated collection framework to facilitate future research in mobile OS navigation.

%% file: review/text/intro.tex
\begin{figure}[t]
    \centering
    \includegraphics[width=\linewidth]{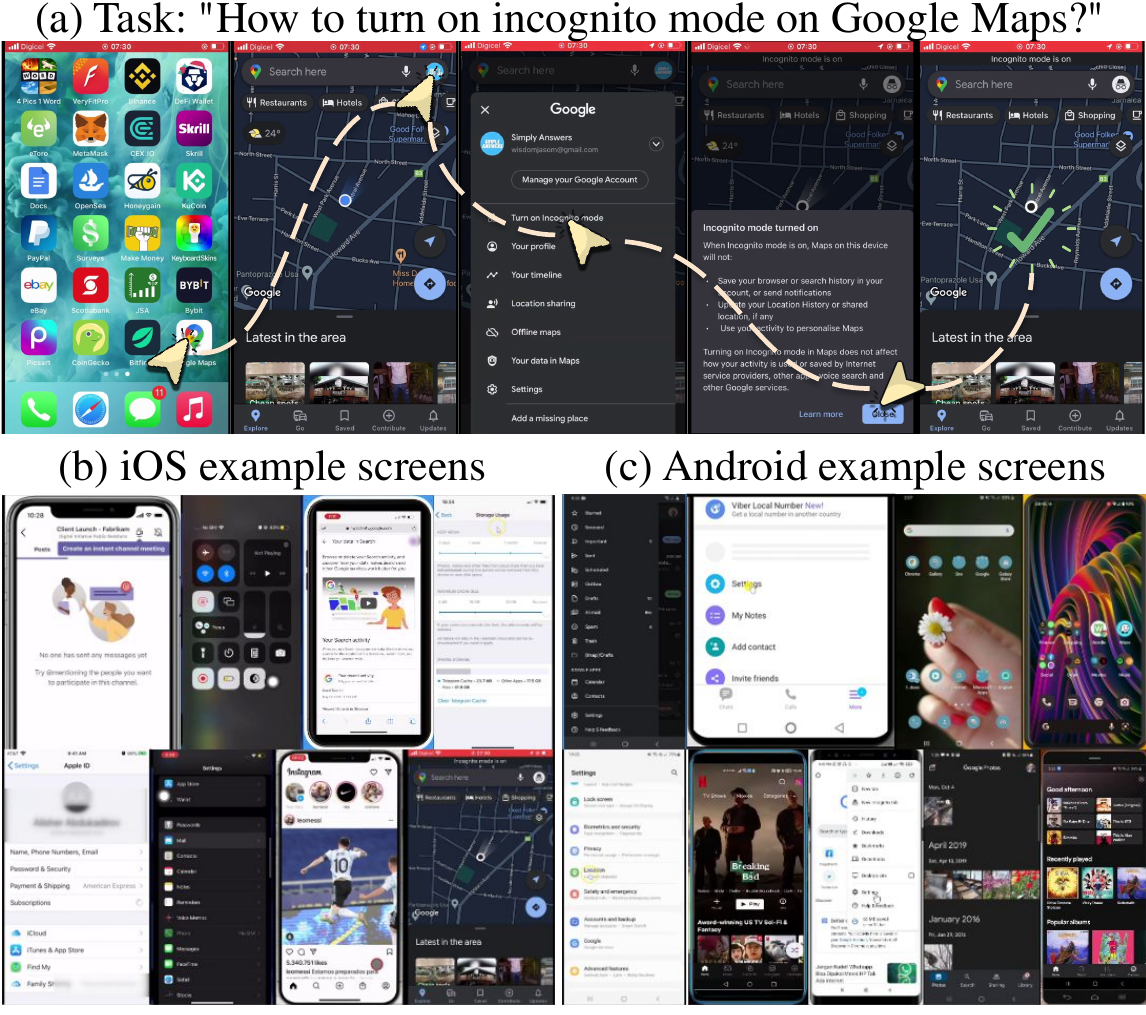}
    \caption{Example screens from our \ourDatasetShortName dataset: (a) mobile OS navigation sequence showing how to turn off incognito mode on Google Maps. Our dataset captures real-world mobile OS navigation procedures across different platforms and configurations, enabling effective generalization; (b) iOS interfaces across different versions and user configurations, including light/dark mode, custom control center, and accessibility settings; (c) Android interfaces with various themes, app layouts, and different resolutions.}
    \label{mobvid:fig:dataset_samples}
\end{figure}

With the rise of Large Language Models (LLMs) \cite{devlin-naacl19,radford-blog19,brown-neurips20,ouyang-neurips22} and Vision-Language Models (VLMs) \cite{liu-neurips23,liu-arxiv23-illava}, agents have increasingly succeeded in executing natural language instructions on GUI systems, including mobile operating systems (OS), using only visual cues \cite{zhang-arxiv23,cheng-acl24}.
These agents enable hands-free device control, which is particularly valuable for users with physical limitations or in situations where manual interaction is impractical.
Such agents also significantly reduce the learning curve for new users while saving time through automated task execution.

A fundamental technical challenge in developing robust mobile OS navigation agents lies in acquiring diverse, real-world training data that reflects the wide variety of UI layouts, elements, and interactions that users encounter in practice.
Existing approaches \cite{zhang-arxiv23,cheng-acl24} rely heavily on datasets that capture device control through \emph{manually annotated} datasets \cite{rawles-neurips23,deka-uist17} or simulated environments \cite{lee-iclrw24,shvo-canai21}.
However, these approaches face several critical limitations: manual annotation is time-consuming, rapid OS updates quickly make existing datasets obsolete, and they cover only a limited range of user configurations and real-world tasks.

To address these challenges, we introduce \ourDatasetShortName (\ourDatasetFullBoldfaceName), a large-scale dataset of 313K annotated frames from 20K instructional videos that captures diverse real-world mobile OS tasks.
Our dataset represents a significant advance in mobile OS navigation, providing unprecedented coverage across different platforms and configurations, as illustrated in Figure \ref{mobvid:fig:dataset_samples}.
We leverage publicly available mobile OS instructional videos on YouTube that contain a rich and abundant range of real-world tasks and environments.
By extracting action sequences from these videos, we build comprehensive and diverse real-world mobile OS task dataset without manual annotation, similar to recent successes in LLMs with web-scale data \cite{devlin-naacl19,radford-blog19,brown-neurips20,ouyang-neurips22}.

To enable continuous dataset expansion as mobile platforms evolve, we present an automated framework that processes instructional videos to create a task dataset.
Our approach comprises robust OCR-based scene detection (95.04\% F1-score), near-perfect UI element detection (99.87\% hit ratio), and a novel multi-step action identification for precise localization.
This automation reliably extracts navigation procedures without requiring platform-specific adaptations.

Moreover, models that include \ourDatasetShortName in their pre-training phases demonstrate superior generalization capabilities across different platforms, while achieving an average performance gain of 18.11\%p on unseen mobile OS compared to existing approaches.
We release our \ourDatasetShortName dataset and models to support future research.

In summary, the key contributions of our work are:
\begin{compactitem}
    \item A large-scale dataset of 313K annotated frames from 20K videos across multiple platforms, created through automated extraction of mobile OS navigation procedures from instructional videos.
    \item A robust OCR-based scene transition detection method that achieves 95.04\% F1-score, outperforming baselines by 12.77\%p across diverse UI configurations.
    \item A novel 3-step action identification method that combines near-perfect UI element detection (99.87\% hit ratio) with temporal reasoning and action localization.
    \item Experimental results demonstrating superior cross-platform generalization, with an average performance gain of 18.11\%p on unseen mobile platform.
\end{compactitem}

This paper is organized as follows: Section \ref{mobvid:sec:related} reviews related work in mobile OS datasets, video instruction mining, and navigation agents.
Section \ref{mobvid:sec:dataset} describes our method of data collection, scene processing, and action annotation.
Section \ref{mobvid:sec:experiments} presents our experimental results and analysis. Finally, Section \ref{mobvid:sec:conclusion} discusses conclusions and future directions.

%% file: review/text/related.tex
\subsection{Mobile OS Datasets and Limitations}
Initial mobile OS datasets have relied on controlled environments and manual annotation. 
While Android in the Wild (AitW) \cite{rawles-neurips23} supports mobile navigation agent, it is limited to Pixel emulators with system logs. 
AndroidControl \cite{li-neurips24}, AMEX \cite{chai-arxiv24} offer Android navigation datasets but lack multi-platform coverage.
Similarly, MobileEnv \cite{zhang-arxiv23-mobileenv} and AndroidEnv \cite{toyama-arxiv21} focus on emulated environments, missing real-world mobile navigation diversity. 
ScreenSpot \cite{cheng-acl24} covers multiple platforms but only supports GUI grounding.

Recent efforts like Video2Action \cite{feng-uist23} and Chen~\etal~\cite{chen-arxiv22} aim to automate task extraction from videos.
Video2Action relies on pixel-level differences, making it unreliable in diverse real-world settings. 
Chen~\etal~'s action identification method requires pre-training on a human-labeled Android dataset \cite{deka-uist17} and struggles with cross-platform generalization.
In contrast, \ourDatasetShortName employs robust scene transition detection and UI element-based action identification to automatically extract navigation procedures without requiring additional training data or simulator environments.

\subsection{Video-based Instruction Mining}
Early video understanding efforts focused on human-annotated datasets like CrossTask \cite{zhukov-cvpr19}, Assembly101 \cite{sener-cvpr22}, and COIN \cite{tang-cvpr19}. 
The costly annotation process led researchers to explore learning from unannotated videos, resulting in datasets like HowTo100M \cite{miech-iccv19} and VLOGs \cite{fouhey-cvpr18}. Recent advancements in using LLMs for video understanding \cite{shang-iccvw23,logeswaran-aclf23,jang-iclrw23} have shown promise in extracting task-specific knowledge from instructional videos, particularly for structured action sequences and temporal relationships. 
However, these approaches primarily focus on physical tasks in real-world environments, while \ourDatasetShortName adapts and extends these techniques specifically for mobile OS.

\begin{table*}[t]
\centering
\begin{tabular}{lccccccc}
\toprule
Dataset & \# Videos & \# Frames & Real-world & Data Access & Code Access & Platforms & Method\\
\midrule
RICO \cite{deka-uist17} & \colorxmark & 72K & \colorxmark & \colorcmark & \colorxmark & Android & Manual \\
AitW \cite{rawles-neurips23} & \colorxmark & 5.7M & \colorxmark & \colorcmark & \colorxmark & Android & Manual  \\
MM-Navigator \cite{yan-arxiv23} & \colorxmark & 110 & \colorcmark & \colorxmark & \colorcmark & iOS + Android & Manual\\
Chen~\etal~\cite{chen-arxiv22} & 128 & 1K & \colorcmark & \colorxmark & \colorxmark & iOS + Android & Manual \\
Video2Action \cite{feng-uist23} & 6K & 30K & \colorcmark & \colorxmark & \colorxmark & Android & Semi-Auto \\
\midrule
\textbf{\ourDatasetShortName (Ours)} & 20K & 313K & \colorcmark & \colorcmark & \colorcmark & iOS + Android & Automated \\
\bottomrule
\end{tabular}
\caption{Comparison of mobile OS navigation datasets. Our dataset provides broader coverage across platforms and configurations while eliminating the need for manual human annotation.} 
\label{mobvid:tab:dataset_comparison}
\end{table*}

\subsection{Mobile OS Agents and Navigation}
Initial navigation agents relied on HTML or system logs for next-action prediction without visual input \cite{deng-neurips24,zhou-iclr23}. 
As the domain expanded and visual cues became essential, multimodal vision-language approaches emerged \cite{koh-acl24,koh-arxiv24,zheng-icml24,niu-arxiv24}.
This shift, along with modern operating systems restricting access to component information, necessitates robust visual understanding and action prediction.
Our work addresses these challenges through a novel 3-step action identification  method for precise cross-platform action localization. 
While recent works have explored reinforcement learning \cite{lee-iclrw24,shvo-canai21} for mobile navigation, their performance is often limited when generalizing to diverse real-world environments. 
To address this gap, we provide a large-scale, diverse dataset of real-world mobile OS navigation procedures that enables  better cross-platform generalization.

%% file: review/text/dataset.tex
We present \ourDatasetShortName (\ourDatasetFullBoldfaceName), a large-scale dataset of mobile OS navigation, captured from real-world instructional videos.
In this section, we first describe the dataset characteristics and key properties, followed by our automated framework for data collection and annotation.

\vspace{0.5em}
\noindent\textbf{Dataset characteristics.}
Our dataset comprises 20K videos with 313K annotated frames, representing a diverse range of mobile OS tasks and navigation procedures.
Our dataset captures a comprehensive range of mobile OS device control, including single-point actions (touch, long press), motion-based actions (scroll, multi touch, zoom in/out), text input (typing), and hardware-specific operations (home, back, volume controls, etc).

The action distribution reflects real-world usage patterns, with touch actions being most frequent at 79.83\%, followed by scroll (8.53\%), hardware interactions (6.73\%), typing (2.68\%), long press (1.11\%), multi touch (0.80\%) and zoom (0.32\%)
A detailed breakdown of action types is provided in Supplementary Section \ref{mobvid:suppsec:pipeline_details}.

Our dataset primarily consists of iOS (49.50\%) and Android (50.50\%) platforms, including both physical devices and simulator recordings.
This diversity in device configurations includes various user settings (\eg, different themes, home screen layouts, accessibility settings) and interface variations that are essential for real-world deployment.

Table \ref{mobvid:tab:dataset_comparison} presents a detailed comparison between \ourDatasetShortName and existing mobile OS datasets.
Unlike AitW \cite{rawles-neurips23} which is based on Android emulator data, our cross-platform support is crucial for developing mobile OS agents with strong generalization.
The automated nature of our collection method eliminates the need for human annotators hired in previous works \cite{deka-uist17,rawles-neurips23,chen-arxiv22,yan-arxiv23} while maintaining high-quality annotations through our multi-step procedures.
Furthermore, this automation allows continuous dataset expansion as new mobile OS versions and features are released, addressing the rapid evolution of mobile interfaces that often renders static datasets obsolete.

The dataset collection process is cost-effective, requiring \$0.34 per video compared to \$5.76 per video for manual annotation by expert annotators (measured from expert annotators on our test set of 100 videos).
This efficiency enables continuous dataset expansion as mobile platforms evolve.
A detailed analysis of computational requirements and costs is provided in Supplementary Section \ref{mobvid:suppsec:additional_dataset_details}.

\begin{figure*}[t]
\centering
\includegraphics[width=\linewidth]{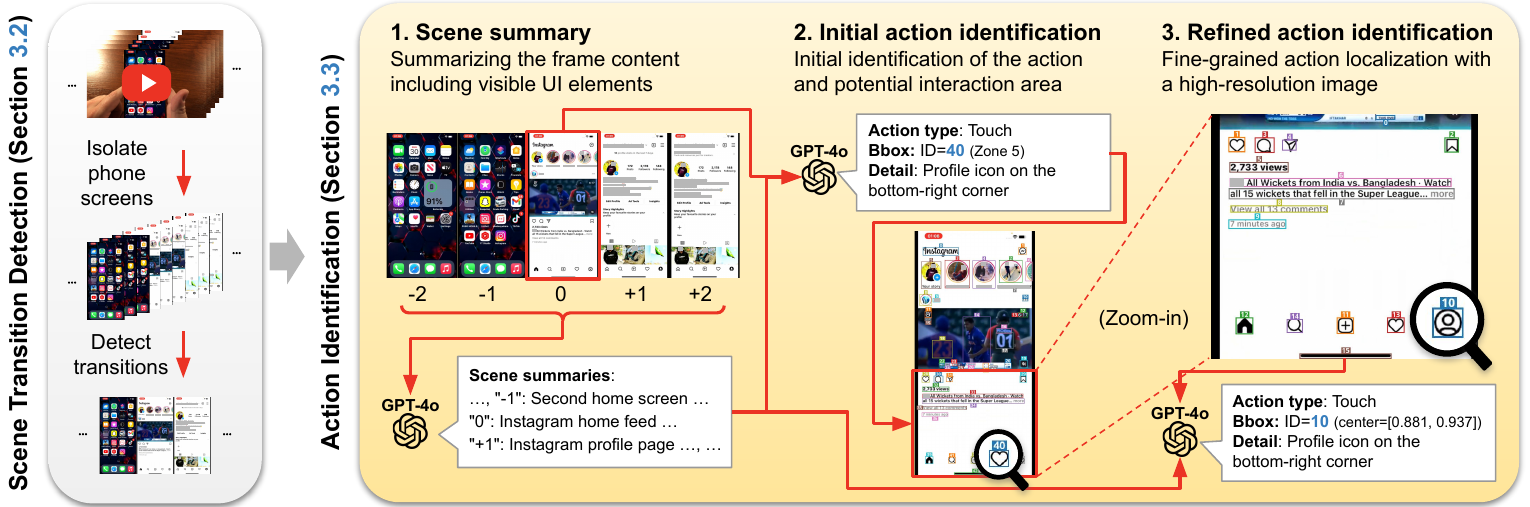}
\caption{\ourDatasetShortName dataset collection framework for mobile OS task dataset for agents from
YouTube. Given a video, we first detect scene transitions (Section \ref{mobvid:subsec:dataset_scene_transition_detection}) and then identify actions in a 3-step process (Section \ref{mobvid:subsec:dataset_action_annotation}): (1) scene summary, (2) initial action identification with SoM representation , and (3) refined action identification for precise localization. In all three steps, we leverage narrations to disambiguate between multiple UI elements of similar effects. The final coordinate is set to the center of the bounding box of the selected UI element.}
\label{mobvid:fig:pipeline}
\end{figure*}

\vspace{0.5em}
\noindent\textbf{Framework overview.}
To create this comprehensive dataset, we developed an automated framework that processes instructional videos to extract mobile OS navigation procedures, as illustrated in Figure \ref{mobvid:fig:pipeline}.
Our goal is to construct a dataset of natural language descriptions of tasks users typically perform on mobile platforms coupled with a sequence of image screenshots (scenes or steps) and corresponding actions \cite{rawles-neurips23,feng-uist23}.
Our framework consists of three main components:

First, we collect and filter instructional videos from YouTube, based on web discussions about mobile OS tasks.
This ensures our dataset captures real-world tasks that users commonly seek guidance for.

Second, we employ OCR-based scene transition detection to identify meaningful scene changes in the mobile OS interface.
This approach proves more robust than traditional vision-based methods across different OS versions and interface configurations.

Third, we combine UI element detection with a novel three-step action identification process.
This includes scene summarization, initial action identification with Set-of-Marks (SoM) representation, and action refinement for precise localization.
The following sections detail each component of our implementation.

\subsection{Mobile Navigation Video Collection}
\label{mobvid:subsec:video_collection}
\noindent\textbf{Task acquisition.}
Our data collection process begins with CommonCrawl web posts, specifically utilizing the C4 \cite{raffel-jmlr20} and Dolma \cite{soldaini-arxiv24} datasets.
These web posts represent actual user discussions and questions about mobile OS tasks, providing a natural distribution of real-world tasks.
This starting point is crucial as it allows us to discover the diverse range of mobile OS tasks that users are interested in, which is not known a priori.
We initially filter web posts using an expanded version of AndroidHowTo's domain whitelist \cite{li-acl20}, which we adapted to include iOS-related websites alongside the original Android domains.
To further refine our selection, we employ GPT 3.5\footnote{\texttt{gpt-3.5-turbo-instruct}} \cite{openai23-gpt-instruct} to filter posts and identify task names that correspond to mobile OS navigation instructions such as ``How to change wallpaper in Android?'' or ``How to turn on the location tag in Instagram?''.

\vspace{0.5em}
\noindent\textbf{Video collection.}
From these filtered posts and their extracted task names, we search and download YouTube videos shorter than 15 minutes that contain English narration transcripts.
We first downloaded 129K videos and retained 20K after our filtering process.
We first use GroundingDINO \cite{liu-eccv24} to filter out videos that do not contain mobile phone screens, retaining 70\% of videos.
For instance, Android Watch or MacOS are filtered out.
After detecting the phone screens, we then filter out the videos that contains scenes where human hands occlude the screen.
Specifically, we use the Google MediaPipe hand landmark detector \cite{lugaresi-arxiv19} to find videos where hand landmarks and mobile phone screens are detected together, keeping 40\% of the remaining videos.
We further filter the video by sampling five frames from the video in an equidistance manner and asking GPT-4o\footnote{\texttt{gpt-4o-2024-08-06} for all GPT-4o throughout this work} \cite{openai24-gpt-4o} to detect the OS and device type from those subsampled frames, preserving 60\% of videos.
We only include Android or iOS mobile phones, as other mobile operating systems comprise less than 1\% of the videos.
This filtering ensures clean, unobstructed views of mobile OS navigation procedures while retaining narrative context through transcripts.
Please refer to Supplementary Section \ref{mobvid:suppsec:video_collection} for further details on our filtering process.

\subsection{Scene Transition Detection}
\label{mobvid:subsec:dataset_scene_transition_detection}

Detecting scene transitions is fundamental to navigation procedure extraction.
A critical challenge lies in identifying meaningful scene transitions: too many intermediate scenes make action identification ambiguous, while skipping important scenes makes the trajectory hard to identify.
Since textual information in mobile interfaces reliably indicates changes in a scene (\eg, page titles, menu items), our Optical Character Recognition (OCR)-based scene transition detection method identifies significant scene transitions by tracking text changes, enabling clearer action trajectories.

\vspace{0.5em}
\noindent\textbf{Isolate phone screens.}
For scene transition detection, we need to identify distinct screen content changes by extracting mobile phone screens from each video.
We detect phone screens at 2 frames per second (FPS) using GroundingDINO \cite{liu-eccv24}, considering that device positions typically do not change rapidly between the transitions.
The detected phone bounding boxes effectively remove distracting backgrounds in real-world videos.
The isolated phone screens serve as our base representation for detecting scene transitions.
During this process, GroundingDINO may occasionally fail to detect the phone screen in some frames, particularly during in-video animations and camera adjustments.
To handle such cases, we apply linear interpolation between successfully detected frames, ensuring continuous phone screen tracking throughout the video.

\vspace{0.5em}
\noindent\textbf{Detect transitions.}
Having isolated the phone screens, we now focus on detecting scene transitions using text content rather than vision-based features (\eg, luminance difference in YUV \cite{feng-uist23}).
While we process phone screen detection at 2 FPS, we increase the frame rate to 4 FPS for OCR analysis since screen content changes occur more frequently than device position changes.
Using Paddle OCR \cite{li-arxiv22}, we extract text and their locations from consecutive frames.
To detect transitions, we track text elements at identical screen locations between adjacent frames, where missing or changed text in subsequent frames is treated as a content change.
We calculate the Levenshtein distance \cite{levenshtein-spd96} between corresponding text elements and mark a transition when the proportion of changed text exceeds 20\% (we empirically set the threshold through our preliminary experiments).
This method proves more effective than vision-based approaches as text rendering remains relatively consistent across different OS versions and user settings (\eg, light/dark mode, recording conditions), as in our evaluation results (Section \ref{mobvid:subsubsec:scene_transition_detection_eval}).
The complete details about our scene transition detection are in Supplementary Section \ref{mobvid:suppsec:pipeline_details}.

\subsection{Action Identification}
\label{mobvid:subsec:dataset_action_annotation}

\subsubsection{UI Element Detection}
\label{mobvid:subsubsec:dataset_ui_component_detection}

Large Vision-Language Models (VLMs) often struggle with precise spatial localization \cite{openai-gpt4v-limitations}.
To address this, we adopt the Set-of-Marks (SoM) approach \cite{yang-arxiv23}, which overlays numbered labels on detected UI elements, when identifying precise action location in the image.
Given the lack of reliable open-source models for UI element detection in mobile interfaces \cite{sunkara-coling22,chen-chi22} or their inferior performance on mobile OS \cite{lu-arxiv24}, we implement a GroundingDINO \cite{liu-eccv24}-based detection module. %
We also obtain the text and its positions from OCR \cite{li-arxiv22}.
We then post-process these detections using mobile-specific heuristic filtering, which merges overlapping bounding boxes and prioritizes UI-appropriate elements based on their shape and relative screen coverage.
The effectiveness of our filtering approach compared to OmniParser \cite{lu-arxiv24} is evaluated in Section \ref{mobvid:subsubsec:ui_component_detection_eval}.
Supplementary Section
\ref{mobvid:suppsec:pipeline_details} provides further details on our approach.

\subsubsection{3-step Action Identification}
\label{mobvid:subsubsec:dataset_three_step_action_prediction}
Mobile interfaces present a particular challenge for action identification as each frame represents a different scene, requiring understanding of both previous and future context to accurately determine the appropriate action.
To address this, our action annotation process employs a novel three-step approach using GPT-4o \cite{openai24-gpt-4o}, incorporating video narration in each step to disambiguate actions in complex scenarios.
Using our SoM representation, we identify actions using numbered labels, which are later converted to screen coordinates using the center points of the bounding boxes of the corresponding UI elements, as shown in Figure \ref{mobvid:fig:pipeline}.

Based on the SoM representation and the narrations, the three steps progressively refine our identified actions as follows.
First, we \textbf{summarize} each frame without UI element markings to provide an unobstructed view of the screen layout and UI elements.
Second, we \textbf{initially identify} a list of actions that can be carried out on the current screen by analyzing the summaries of current and adjacent frames (two previous and two next), along with the SoM representation and narration.
This temporal context helps disambiguate the sequence of actions, while narration provides crucial guidance when multiple UI elements could achieve similar effects.
In the final \textbf{refinement} step, we address VLMs' limitations in precise spatial localization by creating zoomed views around the previously detected UI elements and feeding these views with SoM representation back to GPT-4o.
This zoomed-in approach enables more accurate UI element selection by focusing on specific screen regions.

Figure \ref{mobvid:fig:pipeline} illustrates our complete framework, showcasing how these components work together to extract mobile OS tasks from YouTube videos.
By considering the current frame, adjacent frames, and potential UI elements through this progressive refinement process, our method achieves robust action identification across different platforms and configurations (Section \ref{mobvid:subsubsec:action_annotation_eval}).
With our fully automated framework established, we evaluate models trained on \ourDatasetShortName for mobile OS navigation tasks (Section \ref{mobvid:subsec:dataset_finetuning_performance_evaluation}).

%% file: review/text/experiments.tex
Having established our framework components, we now evaluate both our dataset collection method and models trained on \ourDatasetShortName through comprehensive experiments.

\subsection{Dataset Collection Method Evaluation}
\label{mobvid:subsec:dataset_collection_evaluation}

\subsubsection{Evaluation Dataset}
\label{mobvid:subsubsec:dataset_collection_evaldata}
To evaluate our data collection method, we manually annotated 100 YouTube videos, creating an evaluation dataset of 1,202 frames with 1,070 actions.
Two independent annotators processed each video, with a third resolving disagreements.
Inter-annotator disagreement occurred in only 3.93\% of actions.  %
This test set reflects a representative distribution of real-world mobile OS tasks, with touch actions comprising 67.2\% of all actions, followed by scroll (19.7\%), hardware interactions (7.4\%), typing (3.9\%), zoom (1.0\%) and long press (0.8\%).
The platform distribution maintains 50\% iOS and 50\% Android videos.

\subsubsection{Scene Transition Detection}
\label{mobvid:subsubsec:scene_transition_detection_eval}

Using our evaluation dataset, we first assess scene transition detection performance via F1-score.
We compare our OCR-based approach with two baselines: (a) SceneDetect \cite{pyscenedetect-24} which employs content-aware detection by analyzing frame-by-frame differences through multiple visual features and (b) YUV-diff \cite{feng-uist23} which targets UI transitions by computing luminance differences in YUV colorspace.
As shown in Table \ref{mobvid:tab:scene_transition}, our OCR-based method significantly outperforms baseline approaches, achieving 95.04\% F1-score compared to 82.27\% and 70.86\% for SceneDetect and YUV-diff, respectively.
This performance gap primarily stems from our method's robustness to interface variations, particularly in scenarios where YUV-diff struggles with luminance-based detection in dark mode interfaces, and SceneDetect's approach falls short in capturing the scenes that do not have a clear transition effect.

\begin{table}[t]
\centering
\small
\setlength\tabcolsep{1pt}
\begin{tabular}{@{}lccc@{}}
\toprule
 & YUV-diff \cite{feng-uist23} & SceneDetect \cite{pyscenedetect-24} & OCR-based (\textbf{Ours})   \\
\midrule
F1-score (\%) & 70.86 & 82.27 & \textbf{95.04} \\
\bottomrule
\end{tabular}
\caption{Scene transition detection performance.
Our OCR-based approach significantly outperforms baselines by leveraging text content changes rather than traditional visual features.}
\label{mobvid:tab:scene_transition}
\end{table}
\begin{figure}[t]
    \centering
    \includegraphics[width=\linewidth]{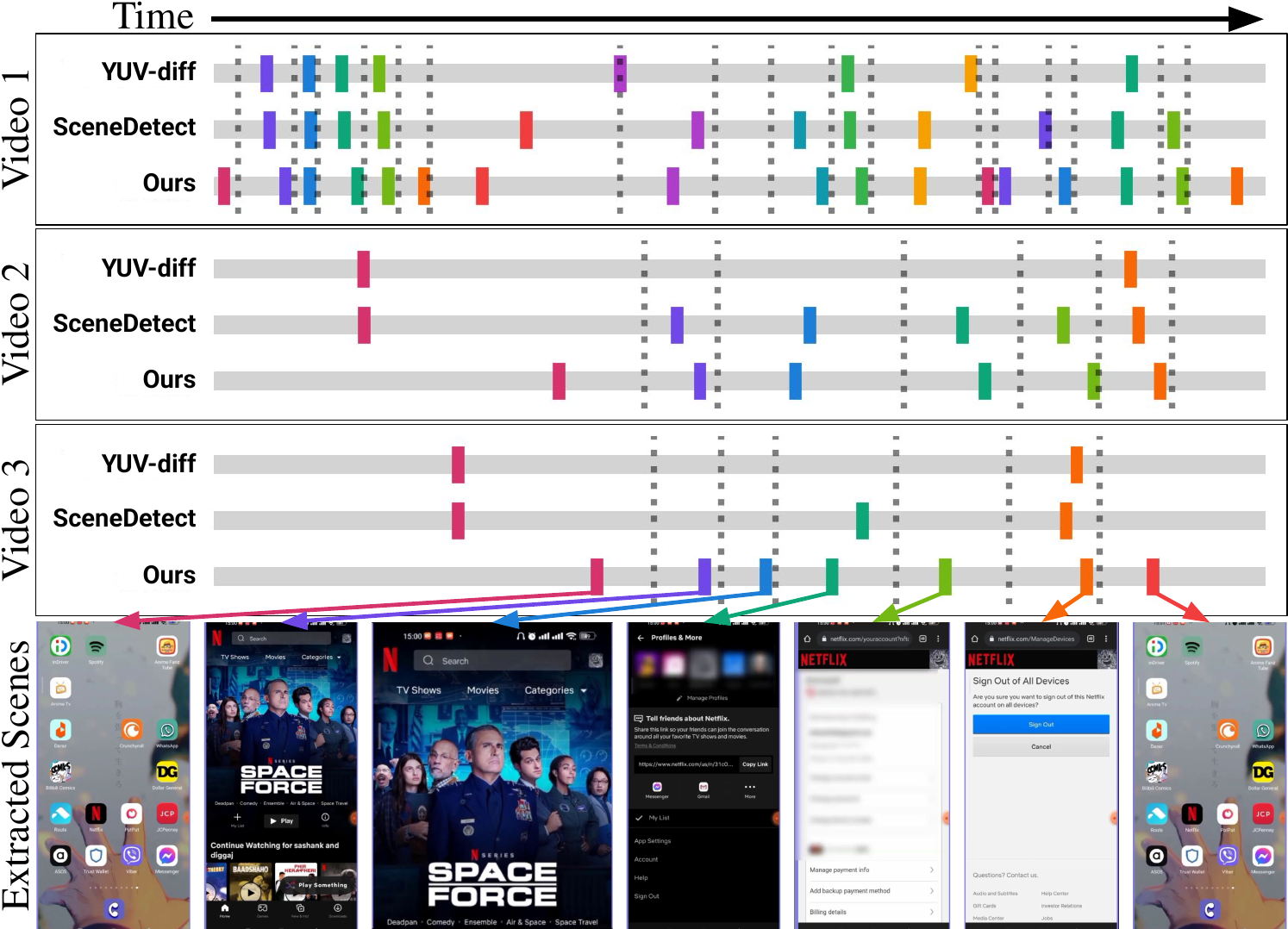}
    \caption{Extracted position of the frame from each scene transition detection.
    Dotted vertical lines represent ground truth transition points, with frames from the same transition segment marked in identical colors.
    Vision-based methods often miss transitions when visual changes are subtle, whereas our OCR-based method reliably detects them.
    }
    \label{mobvid:fig:scene_transition_comparison}
\end{figure}

To further analyze the temporal characteristics of transition detection, we visualize frame selection patterns in Figure \ref{mobvid:fig:scene_transition_comparison}, where dotted vertical lines represent ground truth transition timestamps.
Frames extracted from the same transition segment are marked with identical colors.
Given that mobile OS interfaces maintain stable scene between consecutive transitions, all methods employ a strategic sampling approach: selecting the most representative frame near the temporal midpoint between detected transitions.
This visualization demonstrates how the quality of transition detection directly influences the comprehensiveness of captured scenes, as missed transitions can lead to incomplete task representations in the extracted frame sequence.

\subsubsection{UI Element Detection}
\label{mobvid:subsubsec:ui_component_detection_eval}
We evaluate our UI element detection approach against the recent OmniParser \cite{lu-arxiv24}, with both methods building upon GroundingDINO \cite{liu-eccv24} as their foundation.
While OmniParser creates a general GUI understanding system by fine-tuning their model on web-based datasets and incorporating a separate UI element description model to interpret UI functionality, our method focuses specifically on mobile OS interfaces by combining GroundingDINO with Paddle OCR \cite{li-arxiv22} and carefully designed mobile-specific filtering heuristics.
This targeted approach achieves effective element detection without requiring additional training.

\begin{table}[t]
\small
\setlength\tabcolsep{15pt}
\centering
\begin{tabular}{@{}lcc@{}}
\toprule
 & OmniParser \cite{lu-arxiv24} & \textbf{Ours}  \\
\midrule
Hit Ratio (\%) & 91.83 & \textbf{99.87} \\
\bottomrule
\end{tabular} \\
\caption{UI element detection performance. Please visit Section \ref{mobvid:subsubsec:scene_transition_detection_eval} for the details, including the definition of Hit Ratio.}
\label{mobvid:tab:ui_component_detection}
\end{table}
\begin{figure}[t]
\centering
\includegraphics[width=\linewidth]{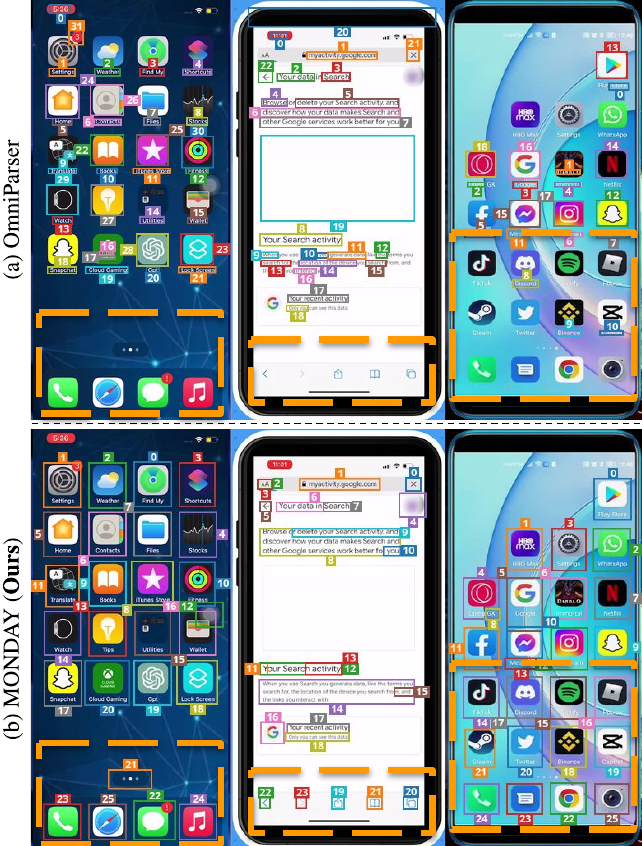}
\caption{Comparison between our UI element detection module and OmniParser \cite{lu-arxiv24}. Ours successfully detects a broader range of UI elements, including home screen icons and bottom-positioned UI elements that OmniParser frequently misses.}
\label{mobvid:fig:ui_component_detection_comparison}
\end{figure}

To evaluate detection performance, we employ Hit Ratio as our primary metric, which assesses whether any detected bounding box's center point falls within the ground-truth interaction area for frames with touch or long press actions.
This metric is particularly relevant as it establishes the upper bound for action identification accuracy through each method's bounding box extraction capability, determining the feasibility of capturing correct interaction points.

Our proposed method demonstrates substantial improvement over OmniParser, achieving a Hit Ratio of 99.87\% on the evaluation dataset.
Quantitatively, our method exhibited exceptional robustness, failing in only a single test case across the entire dataset, while OmniParser demonstrated an error rate of approximately 8\% (Table \ref{mobvid:tab:ui_component_detection}).
Figure \ref{mobvid:fig:ui_component_detection_comparison} provides a qualitative comparison between the bounding boxes generated by our heuristic filter and those produced by OmniParser, with additional examples available in the Supplementary Materials Section \ref{mobvid:suppsec:additional_qualitative_examples}.
The visualization reveals systematic limitations in OmniParser's detection capabilities, particularly in identifying home screen icons and UI elements positioned at the bottom of the screen, while our method successfully detects these UI elements.

\subsubsection{Action Identification}
\label{mobvid:subsubsec:action_annotation_eval}
Following AitW \cite{rawles-neurips23}, we evaluate action identification on two aspects: complete action matching accuracy and touch action matching accuracy.
For touch action, we check whether the identified touch point lies within the correct UI element's bounding box.
Given the lack of code access for previous video-based mobile OS task extraction methods \cite{feng-uist23,chen-arxiv22}, we conducted comprehensive ablation studies to evaluate our multi-step approach:

\begin{compactitem}
    \item 2-step: Omits the final refinement step and uses only scene summary and initial action identification.
    \item 1-step: Direct action identification without intermediate steps (scene summary or refinement).
    \item No narrations: Action identification without narration.
    \item First-step w/ single-image: Uses only the current frame for the first step summary.
\end{compactitem}

Our multi-image 3-step approach consistently outperforms simpler variants across all metrics.
The performance drop from 3-step to 2-step (91.84\% to 89.97\%) in touch operation demonstrates the value of our final refinement stage in precisely localizing actions.
The more substantial decrease to 1-step (74.67\%) shows the complexity of mobile OS tasks and the necessity of multi-stage reasoning.
Further analysis reveals the importance of context: the ``No narrations'' condition shows a significant performance drop (78.20\% vs. 80.84\%), while the single-image approach performs worst (77.22\%), highlighting the importance of both narrative and temporal context in mobile OS navigation.

\begin{figure}[t]
\centering
\vspace{-0.3em}
\includegraphics[width=\linewidth]{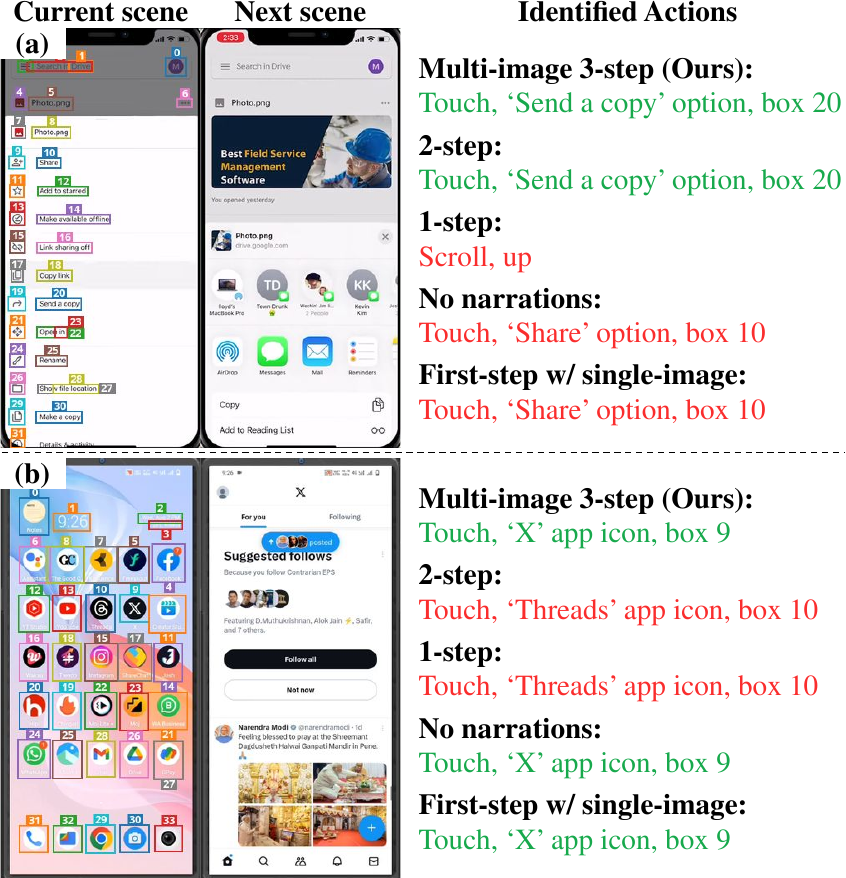}
\caption{Identified actions between different ablation methods. Our multi-image 3-step approach outperforms simplified variants.}
\label{mobvid:fig:action_prediction_comparison}
\end{figure}

Qualitative analysis validates these findings.
As shown in Figure \ref{mobvid:fig:action_prediction_comparison}(a), the model needs both temporal context and narration guidance to select the correct UI element when multiple similar options are available.
Figure \ref{mobvid:fig:action_prediction_comparison}(b) illustrates the superiority of 3-step identification in terms of precise action localization, where 1-step and 2-step methods consistently fail to select the correct UI element.
These visualization results show that our 3-step identification framework, coupled with narration understanding, is essential for robust action identification in complex mobile interfaces.

\begin{table}[t]
\centering
\small
\setlength\tabcolsep{12pt}
\begin{tabular}{@{}lcc@{}}
\toprule
Method & All (\%) & Touch (\%) \\
\midrule
Multi-image 3-step (\textbf{Ours}) & \textbf{80.90} & \textbf{91.84} \\
\midrule
\textbf{Number of steps:} & & \\
\quad 2-step & 79.43 & 89.97 \\
\quad 1-step & 70.63 & 74.67 \\
\midrule
\textbf{Missing context:} & & \\
\quad No narrations & 78.20 & 87.64 \\
\quad First-step w/ single-image  & 77.22 & 89.30 \\
\bottomrule
\end{tabular}
\caption{Action identification accuracy. Our multi-image 3-step approach achieves the best performance between different ablations, demonstrating the importance of each component.}
\label{mobvid:tab:action_prediction}
\end{table}

\subsection{Mobile Navigation Agent Evaluation}
\label{mobvid:subsec:dataset_finetuning_performance_evaluation}

Having established the effectiveness of our framework components, we now evaluate the impact of incorporating \ourDatasetShortName into the model pre-training phase on downstream mobile navigation performance.
For a comprehensive evaluation of the impact, we contrast each baseline pre-trained model with its \ourDatasetShortName-induced variant on 4 different test sets, AitW \cite{rawles-neurips23}, AMEX \cite{chai-arxiv24}, \ourDatasetShortName, and Windows Mobile (unseen mobile platform), under 2 separate finetuning scenarios on AitW and AMEX.

\subsubsection{Baselines and Experimental Setup}
\label{mobvid:subsubsec:finetuning_baseline_and_setup}

We compare each vision-language model with its corresponding \ourDatasetShortName-induced variant, which is obtained by finetuning the model on \ourDatasetShortName via LoRA \cite{hu-iclr22}.
These variants, referred to as SeeClick-\ourDatasetShortName and Llama-3.2-\ourDatasetShortName, are based on SeeClick \cite{cheng-acl24}, which builds on Qwen-VL \cite{bai-arxiv23} with GUI-specific grounding, and Llama-3.2-11B-Vision-Instruct \cite{meta-llama32}, a large-scale model trained on 6 billion image-text pairs.
To ensure a fair comparison, all models are further finetuned using LoRA on either the AitW or AMEX dataset for an equal number of epochs, and the checkpoint with the lowest validation loss is selected for testing.
During finetuning, each model receives the current screen, task name, and the last four actions as input, and predicts the next action.

Following AitW's evaluation protocol \cite{rawles-neurips23}, we measure exact action matching between predictions and ground truth from AitW, AMEX, and \ourDatasetShortName test sets.
For touch and long press actions, matches are validated against annotated interaction regions.
For a fair comparison across datasets, we restrict evaluation to actions common to all datasets.

We further test on an entirely unseen mobile platform, Windows Mobile OS, to evaluate generalization capabilities.
Following the same annotation protocol used in the previous section, we manually annotated 50 videos containing 605 valid frames and 554 actions.
This presents a significant challenge to all finetuned models as Windows Mobile employs distinct UI patterns and interaction paradigms not present in iOS and Android.
Please visit Supplementary Section \ref{mobvid:suppsec:model_finetuning} for the details about experiment settings.

\subsubsection{Results and Analysis}
\label{mobvid:subsubsec:finetuning_results}

Table \ref{mobvid:tab:cross_platform} presents model performance across different test sets.
For AitW, we report average performance over five categories (Google Apps, General, Web Shopping, Install, and Single; see Supplementary Section \ref{mobvid:suppsec:model_finetuning} for per-category performance).
Notably, the models finetuned from \ourDatasetShortName-induced pre-trained models mostly perform better on the AitW and AMEX test sets, while achieving substantially higher performance on the \ourDatasetShortName test set.

The performance gap between models finetuned from the original pre-trained models (SeeClick, Llama-3.2) and the corresponding \ourDatasetShortName-induced variants as the base models highlights the importance of learning from real-world usage patterns.
While AitW and AMEX provides valuable training data, simulator environments cannot fully capture the diversity of real-world deployments, including various task types and user-specific configurations.
We believe models trained on our dataset can handle this simulation-to-real domain gap more effectively.

Moreover, the models finetuned from the \ourDatasetShortName-induced pre-trained models significantly outperform the baselines with the original pre-trained models on the unseen platform (Windows Mobile), as shown in Table \ref{mobvid:tab:cross_platform}.
We believe this successful generalization can be attributed to several key factors.
First, our dataset's multi-platform nature and diversity help models learn platform-agnostic navigation patterns.
Second, exposure to various UI layouts, themes, and custom settings enables better adaptation to novel interfaces.
Finally, the breadth of real-world data captures authentic device controls across OS versions and configurations, allowing models to develop robust navigation capabilities that transfer effectively to unseen platforms.

\begin{table}[t]
\centering
\setlength\tabcolsep{2.3pt}
\footnotesize
\begin{tabular}{@{}lcccc@{}}
\toprule
\multirow{2}{*}{Finetuned model}
& &&Test set \\ \cline{2-5} \noalign{\vskip 1mm}
& AitW & AMEX & \ourDatasetShortName & Windows Mobile \\
\midrule
\multicolumn{4}{l}{{\textbf{\textit{AitW-finetuned from}}:}} \\
\quad SeeClick	     				&	66.98			&	47.23			&	40.66			&	38.54 \\
\quad SeeClick-\ourDatasetShortName	&	\textbf{68.47}	&	\textbf{47.76}	&	\textbf{63.39}	&	\textbf{51.71} \\
\specialrule{0.1pt}{0.5pt}{1.5pt}
\multicolumn{4}{l}{\textbf{\textit{AMEX-finetuned from}}:} \\
\quad SeeClick	     				&	37.08			&	\textbf{68.19}	&	44.23			&	43.17 \\
\quad SeeClick-\ourDatasetShortName	&	\textbf{40.19}	&	66.13			&	\textbf{63.39}	&	\textbf{55.37} \\
\midrule
\multicolumn{4}{l}{\textbf{\textit{AitW-finetuned from}}:} \\
\quad Llama-3.2	     				&	58.96			&	43.74			&	39.80			&	26.83 \\
\quad Llama-3.2-\ourDatasetShortName  &	\textbf{67.38}	&	\textbf{55.96}	&	\textbf{57.99}	&	\textbf{50.24} \\
\specialrule{0.1pt}{0.5pt}{1.5pt}
\multicolumn{4}{l}{\textbf{\textit{AMEX-finetuned from}}:} \\
\quad Llama-3.2	     				&	29.81			&	61.30			&	40.17			&	28.29 \\
\quad Llama-3.2-\ourDatasetShortName  &	\textbf{42.96}	&	\textbf{72.36}	&	\textbf{58.35}	&	\textbf{51.46} \\
\bottomrule
\end{tabular}
\caption{Comparison of navigation action accuracies with the original pre-trained models (SeeClick, Llama-3.2) vs. the corresponding \ourDatasetShortName-induced variants (SeeClick-\ourDatasetShortName, Llama-3.2-\ourDatasetShortName).
Performance on AitW test set is averaged across its evaluation categories.
Models finetuned from \ourDatasetShortName-induced variants mostly outperform the baselines and generalize well to an unseen mobile platform (Windows Mobile).
}
\label{mobvid:tab:cross_platform}
\end{table}

%% file: review/text/conclusion.tex
We presented \ourDatasetShortName, a novel approach for automatically extracting mobile OS navigation procedures from instructional videos. 
Our method eliminates the need for manual annotation through a carefully designed framework combining OCR-based scene detection, robust UI element identification, and multi-step action identification. 
Experiments demonstrate that models trained on our dataset achieve superior generalization across platforms and significantly better adaptation to unseen mobile OS interfaces.

While our current implementation relies on GPT-4o \cite{openai24-gpt-4o} for action identification, the framework's effectiveness in extracting accurate action sequences without human intervention represents an important step toward scalable mobile OS navigation datasets. 
The modular design allows for integration of specialized models, or replacing GPT-4o, as they become available, making the system adaptable to future improvements in model capabilities.

We believe this work opens new possibilities for developing more robust and adaptable GUI visual agents for mobile OS, particularly for real-world applications where interface diversity and cross-platform operation are essential. 
Organizations can apply this approach to their own instructional videos, enabling continuous adaptation to new interface patterns and OS versions as mobile platforms evolve.

%% file: review/supp/additional_dataset_details.tex
\begin{table}[t]
\centering
\centering
\setlength\tabcolsep{15pt}
\begin{tabular}{c|rr|r}
\toprule
         & iOS    & Android & Total  \\
\midrule
Train    & 9,755 & 9,970  & 19,725 \\
Val      & 246    & 249     & 495    \\
Test     & 50     & 50      & 100    \\
\midrule
Total    & 10,051 & 10,269  & 20,320 \\
\bottomrule
\end{tabular}
\caption{Distribution of videos across different splits in \ourDatasetShortName. Validation and test sets are manually balanced between platforms, while training set maintains natural distribution from collection process.}
\label{mobvid:supptab:dataset_dist}
\end{table}
\begin{figure}[t]
    \centering
    \includegraphics[width=\linewidth]{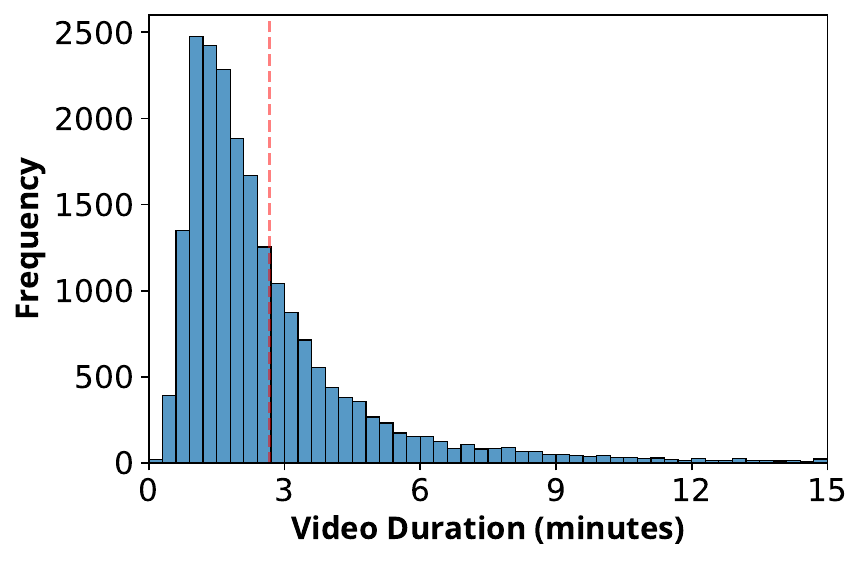}
    \caption{Distribution of video duration in minutes. Red vertical dotted line stands for the average duration of 2.66 minutes. The majority of videos (77.8\%) fall between 1-5.5 minutes, with a peak at 1.05 minutes.}
    \label{mobvid:suppfig:video_duration}
\end{figure}

\begin{figure}[t]
    \centering
    \includegraphics[width=\linewidth]{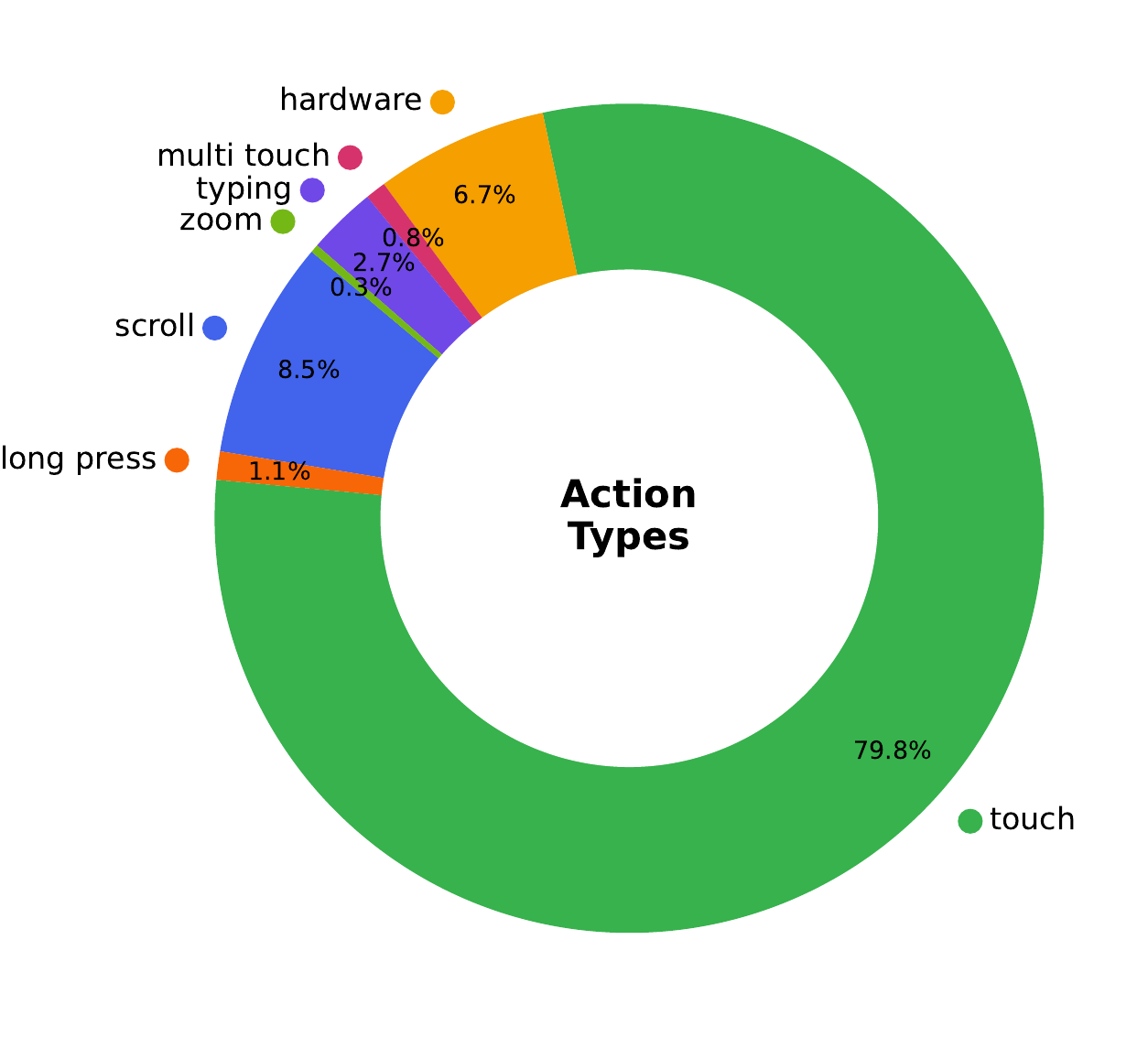}
    \caption{Action type distribution in our dataset shows touch actions dominate at 79.83\%, followed by scroll (8.53\%) and other actions.}
    \label{mobvid:suppfig:action_dist}
\end{figure}

\subsection{Dataset Distribution}
Our dataset is split into 19,725 training videos, 495 validation videos, and 100 test videos, as shown in Table \ref{mobvid:supptab:dataset_dist}. 
The validation set contains an equal distribution of 246 iOS and 249 Android videos, while the test set maintains the same balanced 50/50 split between platforms. 
The training set includes 9,755 iOS and 9,970 Android videos, reflecting the natural distribution from our collection process.

As shown in Figure \ref{mobvid:suppfig:video_duration}, our dataset primarily consists of concise, focused instructional videos with an average duration of 2.66 minutes. 
The duration distribution shows a clear peak at 1.05 minutes, with 77.8\% of videos falling between 1-5.5 minutes. 
This distribution reflects the typical length of mobile OS instructional content, which tends to focus on specific, well-defined tasks.

The distribution of actions in our dataset reflects real-world usage patterns, as illustrated in Figure \ref{mobvid:suppfig:action_dist}. 
Touch actions are the most frequent (79.83\%), followed by scroll (8.53\%), hardware interactions (6.73\%), typing (2.68\%), long press (1.11\%), multi touch (0.80\%) and zoom (0.32\%).

In terms of app coverage, we checked which mobile app each video of MONDAY is for: it includes 2,479 unique apps across 20,337 videos. 
The distribution between OS native and third-party apps (37.6\% : 62.4\%) demonstrates balanced representation of mobile device usage.
Third-party app usage aligns with real-world scenarios, as shown by top applications: Instagram (3.72\%), Facebook (2.60\%), YouTube (2.08\%), Twitter (1.86\%), WhatsApp (1.75\%), and so on.

\subsection{Computational Cost Analysis}

Our framework's processing time is proportional to the inference time of its core components: one Paddle OCR inference and two GroundingDINO inferences (for phone screen and icon detection) per frame, plus three GPT-4o queries per action identification. 
For a typical three minute video, the total processing time prior to the GPT-4o is approximately 9.7 minutes on a single NVIDIA Titan Xp GPU. 
The total cost for identifying actions for the 20,320 videos with GPT-4o was \$6976, approximately \$0.34 per video. %

To better compare its effectiveness, we measure the cost when we ask the annotator to annotate the scene detection and action identification for 100 test videos used in Section \ref{mobvid:subsubsec:dataset_collection_evaldata}. 
Scene transition and action annotation takes 12 minutes and costs \$5.76 per video on average from an expert annotator.  %
If there are good open-source models that perform reasonably well for action identification, then we can further reduce the cost by replacing GPT-4o with these alternatives.

%% file: review/supp/video_collection.tex
\begin{figure}[t]
\centering
\includegraphics[width=\linewidth]{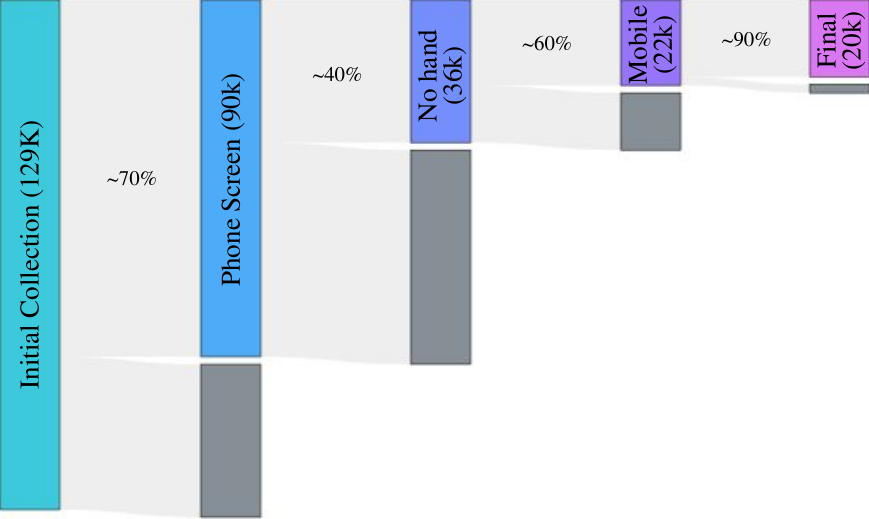}
\caption{Filtering stages in video collection process. Starting from 129K YouTube videos with English transcripts and duration under 15 minutes, each stage progressively filters videos to ensure quality and relevance.}
\label{mobvid:suppfig:video_filtering}
\end{figure}

Our dataset collection process starts by mining mobile OS-related content from CommonCrawl web posts in the C4 \cite{raffel-jmlr20} and Dolma \cite{soldaini-arxiv24} datasets. 
To ensure the mobile OS navigation topic, we first filter these posts using an expanded version of AndroidHowTo's domain whitelist \cite{li-acl20}, which we augmented to include iOS-related websites alongside the original Android domains.
We then employ GPT-3.5 Turbo Instruct \cite{openai23-gpt-instruct} to analyze the main body text of each filtered post, identifying titles that describe iOS/Android phone navigation tasks (responding with ``N/A'' for irrelevant content).
These extracted titles are then used as search keywords for collecting relevant YouTube videos.

From our initial collection of 129K videos that have English transcripts and are shorter than 15 minutes, we implement a multi-stage filtering process, as shown in Figure \ref{mobvid:suppfig:video_filtering}:
\begin{compactitem}
    \item Process videos at 2 FPS using GroundingDINO, requiring successful detection in at least for 30 seconds (retaining 70\% of videos)
    \item Process videos using Google MediaPipe hand landmark detection and filter out videos where hands appear, ensuring clean views of the interface (keeping 40\% of remaining videos)
    \item Sample 5 frames in equidistance and ask GPT-4o to determine OS type (`iOS', `Android', `Windows Mobile', `BlackBerry OS', `Multiple OS', or `None') and device type (`Phone', `Tablet/Pad', `Watch', `Laptop', `Multi-device', or `None'), preserving 60\% of videos
    \item Remove videos with more than 55 detected scenes to ensure focused, single-task demonstrations\
    \item After sampling the evaluation dataset, remove contaminated videos identified by an n-character overlap \cite{openai-arxiv23} (n=30) in video titles
\end{compactitem}

This multi-stage filtering process results in our final dataset of 20K videos capturing clear, unobstructed mobile OS navigation procedures while retaining narrative context through transcripts.

%% file: review/supp/pipeline_details.tex
Our framework, illustrated in Figure \ref{mobvid:fig:pipeline}, consists of three main components working together to extract mobile OS navigation procedures from instructional videos. 
The framework begins with scene transition detection (Section \ref{mobvid:subsec:dataset_scene_transition_detection}), which identifies meaningful state changes in the mobile interface using OCR-based analysis. 
This is followed by UI element detection (Section \ref{mobvid:subsubsec:dataset_ui_component_detection}), which combines icon detection and text recognition to identify interactive elements. 
Finally, our three-step action identification process (Section \ref{mobvid:subsubsec:dataset_three_step_action_prediction}) leverages these detected components along with temporal context and to determine precise user actions.
We will release our complete framework implementation upon acceptance to facilitate future research in mobile OS navigation.

\subsection{Scene Transition Detection}
For phone screen detection, we use GroundingDINO \cite{liu-eccv24} for all frames in 2 FPS with the following parameters:
\begin{compactitem}
    \item Box confidence threshold: 0.25
    \item Text confidence threshold: 0.25
    \item Caption prompt: ``phone screen''
\end{compactitem}
During this process, GroundingDINO may occasionally fail to detect the phone screen in some frames, particularly during in-video animations and camera adjustments. 
To handle such cases, we apply linear interpolation between successfully detected frames within a 3-second window, ensuring continuous phone screen tracking throughout the video.

After detecting the phone screens, our OCR-based scene transition detection algorithm operates as follows:
\begin{compactitem}
    \item Extract text from consecutive frames in 4 FPS using Paddle OCR \cite{li-arxiv22}
    \item Compute the Levenshtein distance \cite{levenshtein-spd96} between the text in an identical location but in adjacent frames
    \item Mark as transition if the distance exceeds 20\% of the number of original text characters
\end{compactitem}

We apply several refinements to ensure robust transition detection:
\begin{compactitem}
    \item Filter OCR results by confidence score ($>$ 0.9) to focus on reliable text detections
    \item Ignore text detected in top 5\% and bottom 10\% of the screen to avoid system-specific UI elements
    \item Merge transitions occurring within 0.4 seconds to handle animation effects
    \item Consider temporal context up to 2 seconds before and after each potential transition for verification
    \item Apply text normalization using regular expressions to handle minor rendering variations
\end{compactitem}

When multiple transitions are detected in close proximity, we select the most representative frame for each transition segment, typically choosing the frame closest to the temporal midpoint between transitions. 
This approach helps capture stable states while filtering out intermediate animation frames.

\subsection{UI Element Detection}
Our UI element detection combines icon detection using GroundingDINO and text detection using OCR, followed by careful filtering to identify genuine interactive elements. 
The system employs a two-stage approach.

First, we detect potential UI elements using GroundingDINO with relaxed thresholds:
\begin{compactitem}
    \item Box confidence threshold: 0.04  %
    \item Text confidence threshold: 0.25
    \item Caption prompt: ``icon''
\end{compactitem}
We deliberately use a lower box confidence threshold here to maximize UI element detection coverage, relying on our subsequent filtering steps to remove false positives.

Then, we apply mobile-specific filtering heuristics:
\begin{compactitem}
    \item Integrate OCR-detected text element boxes
    \item Remove oversized elements (box area $>$ 0.4 of screen)
    \item Merge overlapping boxes with significant intersection (IoU $>$ 0.5)
    \item Filter by aspect ratio and relative positioning
\end{compactitem}

For text elements, we perform additional processing to identify interactive text components like context menu options (\eg, `more' button in text posts) or actionable labels (\eg, `unsubscribe' button in emails):
\begin{compactitem}
    \item Split text by natural spaces
    \item Compute box for each text segment, split by a white space, based on character count
    \item Set dominant color as background
    \item Select next dominant color as text color
    \item Add box if color difference in LAB space $>$ 50 (with step-wise reduction by 5 until text box detection succeeds)
\end{compactitem}

\subsection{Action Identification}
Our action identification process follows a three-step approach to ensure accurate action prediction:

1. Scene Summary: First, we analyze each frame independently to understand the overall UI layout and component relationships, creating a comprehensive scene description without any preconceptions about actions.

2. Initial Action Identification: Using the scene summaries and temporal context from adjacent frames, we identify potential actions that could lead to the observed state changes, considering both visible UI elements and narrative guidance.

3. Refined Action Identification: Finally, we employ a zone-based system for precise spatial localization of the predicted action, dividing the screen into five vertical zones based on UI element positions.
Zones are calculated as follows:
\begin{compactitem}
    \item Zone 1: ~~0.0 - ~45.0\% of screen height (top)
    \item Zone 2: 12.5 - ~57.5\% of screen height
    \item Zone 3: 25.0 - ~70.0\% of screen height
    \item Zone 4: 37.5 - ~82.5\% of screen height
    \item Zone 5: 55.0 - 100.0\% of screen height (bottom)
\end{compactitem}

As a result of three step identification, \ourDatasetShortName captures the following categories of mobile OS device control:
\begin{compactitem}
\item \textbf{Single-point actions:}
  \begin{compactitem}
  \item touch: Single tap at specific coordinates
  \item long press: Extended press at specific coordinates
  \end{compactitem}

\item \textbf{Motion-based actions:}
  \begin{compactitem}
  \item scroll: [up, down, left, right]
  \item zoom: [in, out]
  \item multi touch: swipe (up/left/right), four-finger pinch, double tap, rotate content (clockwise/counterclockwise), multi taps
  \end{compactitem}

\item \textbf{Hardware interactions:}
  \begin{compactitem}
  \item Navigation: home, recent apps (Android-only), back double/triple taps
  \item Device controls: volume up/down, power, authentication
  \item Physical actions: shake, orientation change (clockwise/counterclockwise), silent mode change on/off
  \end{compactitem}

\item \textbf{Text input:} Typing actions with corresponding text content
\end{compactitem}

%% file: review/supp/evaluation_dataset_collection.tex
We employed two experienced annotators familiar with both iOS and Android platforms for evaluation dataset. 
The annotation process consisted of two main tasks:

\textbf{Scene Transition detection.} 
Annotators identified transition points in videos, with timestamps aligned between annotators using minimum distance matching. 
When transition counts differed between annotators, a third annotator reviewed the unmatched timestamps to determine the correct transitions.

\textbf{Action identification.} 

\begin{figure*}[ht!]
\begin{minipage}{\textwidth}
\begin{lstlisting}[language=xml, 
    basicstyle=\small\ttfamily,
    stringstyle=\color{purple},         %
    identifierstyle=\color{red},        %
    showstringspaces=false,
    captionpos=b,
    caption={Label Studio interface configuration for action annotation.},
    label={mobvid:supplst:label_studio}]
<View>
  <Header value="File: $image"/>
  <RectangleLabels name="label" toName="image" fillOpacity="0.7" strokeWidth="3">
    <Label value="click" background="blue"/>
    <Label value="long_press" background="red"/>
  </RectangleLabels>
  <TextArea name="typing" toName="image" editable="true" required="false" 
            maxSubmissions="1" placeholder="typed_text"/>
  <Image name="image" value="$image"/>
  <Choices name="other_actions" toName="image" choice="multiple">
    <Choice alias="end_of_video" value="End of the video"/>
    <Choice alias="ambiguous" value="Ambiguous"/>
    <Choice alias="hardware_recentapps" value="Hardware - Recent Apps (Android left key)"/>
    <Choice alias="hardware_home" value="Hardware - Home"/>
    <Choice alias="hardware_back" value="Hardware - Back (Android right key)"/>
    <Choice alias="hardware_authentication" value="Hardware - Authentication"/>
    [Additional action choices...]
  </Choices>
</View>
\end{lstlisting}
\end{minipage}
\end{figure*}

Using our scene transition detection output, annotators labeled actions between consecutive scenes using Label Studio with a custom interface. 
The annotation interface supported the layout in Listings \ref{mobvid:supplst:label_studio}.

When cases were ambiguous (no clear single action between scenes), annotators marked them as `ambiguous' and these were excluded from evaluation. 
For any disagreements between annotators, a third annotator made the final decision.

Annotation was conducted at a rate of \$16/hour, with each annotator spending approximately 6 hours on scene transition detection and 7 hours on action identification. 
The presence of ground-truth video and annotator expertise in both platforms contributed to high initial agreement rates.
We followed this exact same annotation protocol and quality control process when creating our Windows Mobile test set of 50 videos, ensuring consistent evaluation criteria across all platforms.

%% file: review/supp/additional_qualitative_examples.tex
\begin{figure}[t]
\centering
\includegraphics[width=\linewidth]{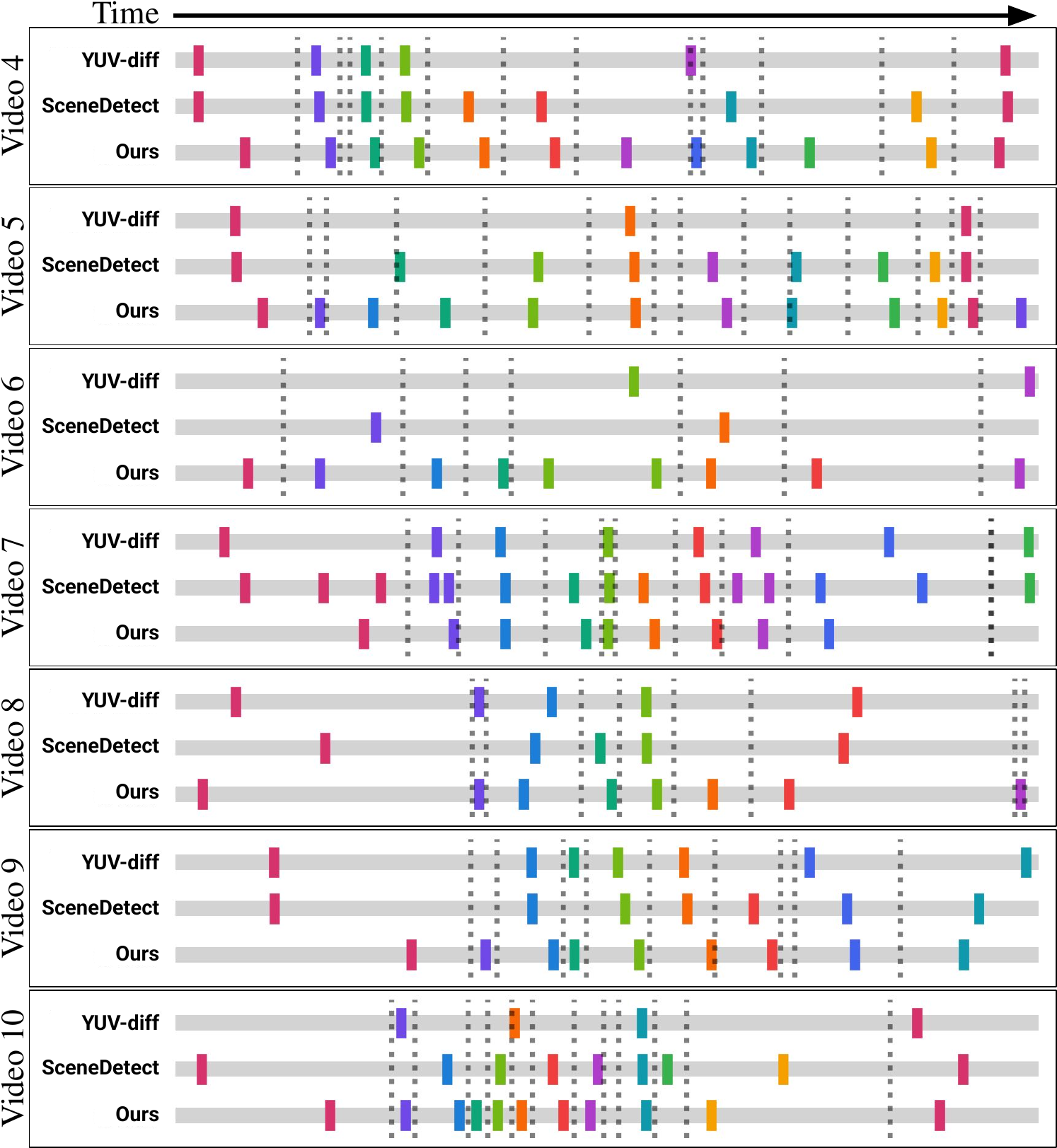}
\caption{Additional examples of scene transition detection results across different interface configurations. Our OCR-based method successfully handles most transitions, though it missed one segment in Video 5 and detected two segments in Video 6. Even with these edge cases, our approach achieves more accurate transition detection compared to baseline methods. See Section \ref{mobvid:subsubsec:scene_transition_detection_eval} for detailed experimental settings.}
\label{mobvid:suppfig:scene_transition}
\end{figure}
\begin{figure*}[t]
\centering
\includegraphics[width=\linewidth]{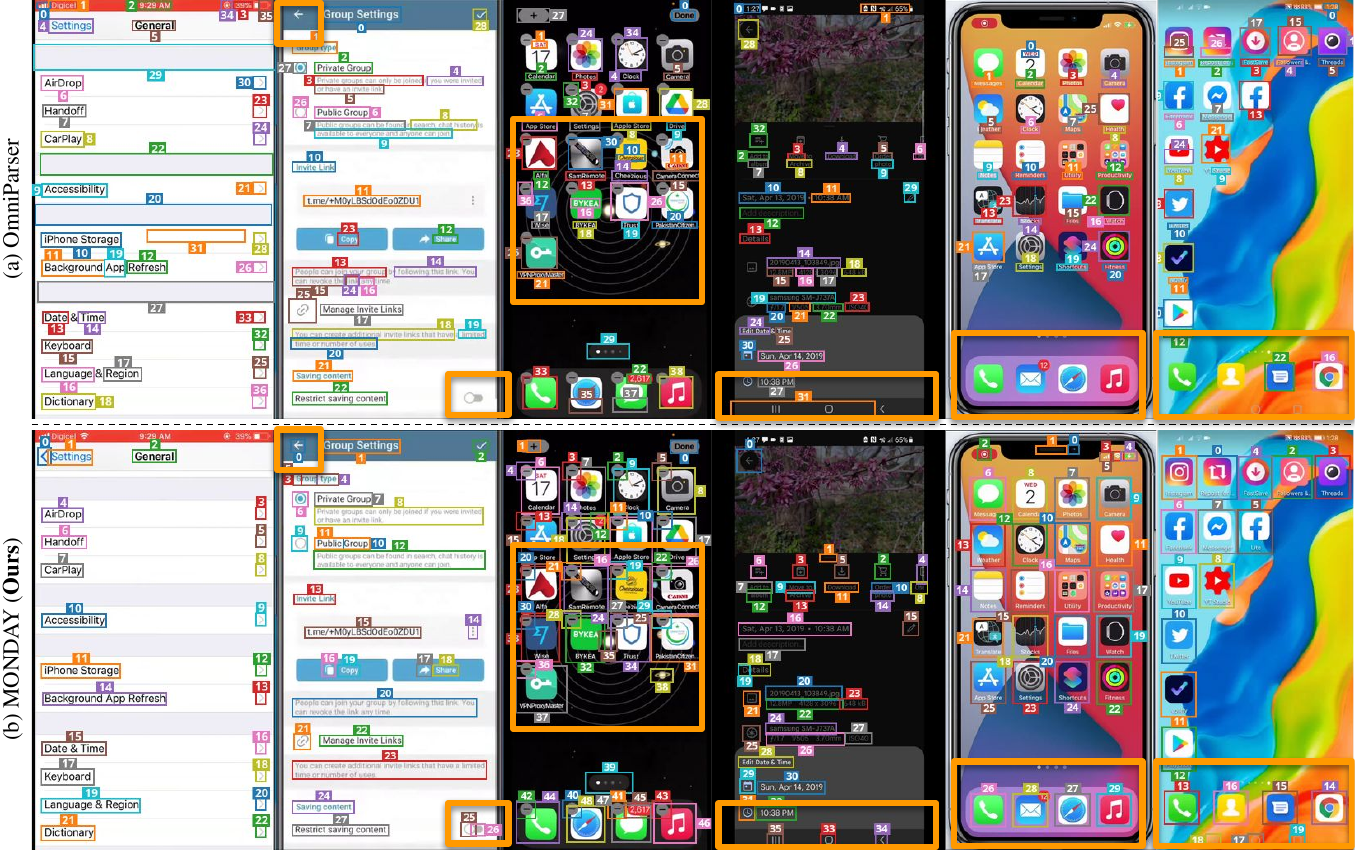}
\caption{Additional comparisons between (a) OmniParser \cite{lu-arxiv24} and (b) our UI element detection module. While OmniParser detects more boxes in the first column, many are not interactable elements. In the next three columns, OmniParser fails to detect important actionable UI elements (\eg, back button, delete button). The last two columns show OmniParser's consistent failure to detect the lower portions of home screen icons.}
\label{mobvid:suppfig:ui_detection}
\end{figure*}
\begin{figure*}[t]
\centering
\includegraphics[width=\linewidth]{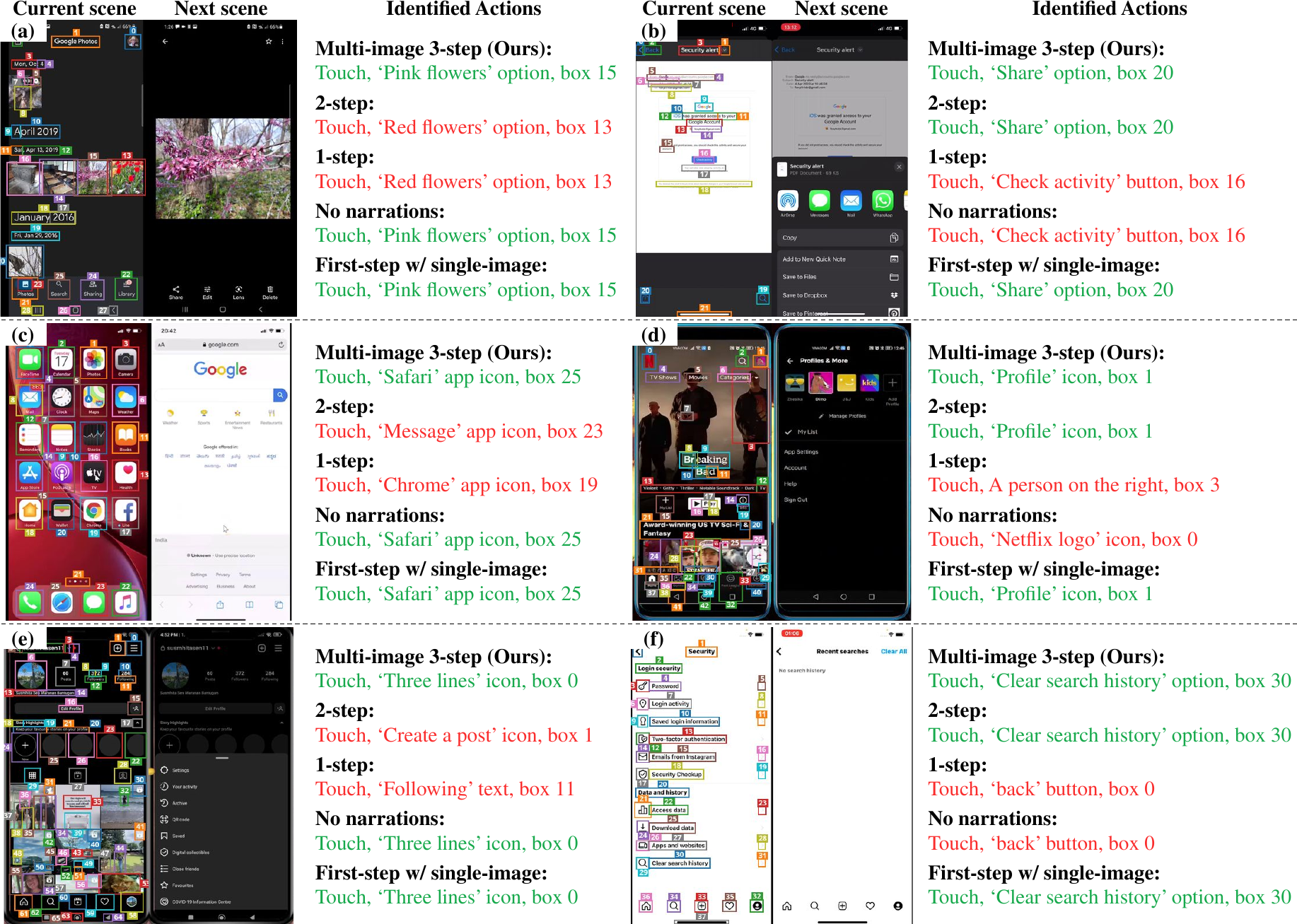}
\caption{Additional comparisons of action identification results between different approaches. The examples highlight two common types of errors: (a,c,e) selecting nearby but incorrect UI elements, as shown in the left column examples, and (b,d,f) cases requiring complex reasoning with audio transcription (ASR) for correct identification, as demonstrated in the right column examples.}
\label{mobvid:suppfig:action_identification}
\end{figure*}

In this section, we provide additional examples demonstrating the effectiveness of our framework components.
Figure \ref{mobvid:suppfig:scene_transition} shows extended cases where our OCR-based scene transition detection successfully handles challenging scenarios.
Figure \ref{mobvid:suppfig:ui_detection} illustrates our UI element detection system's ability to handle complex interface layouts.
Figure \ref{mobvid:suppfig:action_identification} presents comparisons between our multi-step action identification approach and simpler variants.

%% file: review/supp/human_evaluation.tex
We conducted a human evaluation involving 10 workers examining 100 randomly sampled sequences in \ourDatasetShortName training set, with each sequence reviewed by two people. 
The evaluators assessed whether the identified action is accurate, inaccurate, or not enough information to answer, based on the current, two previous and two next scenes with the title.
Workers found that 80.40\% of 250 sampled actions were accurate, while `not enough information (8.60\%)' primarily stemmed from either insufficient context window coverage or incomplete user configuration details. 
This human evaluation, along with our model's test accuracy of 80.90\%, indicates inherent task complexity due to incomplete context and interface ambiguity.

%% file: review/supp/model_finetuning.tex
\subsection{Training Details}

We apply LoRA finetuning \cite{hu-iclr22} for all models using the PEFT library \cite{peft} with its default configuration on their public repository: $\text{lora}_\alpha = 16, \; \text{lora}_r = 64, \; \text{lora}_\text{dropout} = 0.05$ for SeeClick, and $\text{lora}_\alpha = 32, \; \text{lora}_r = 8, \; \text{lora}_\text{dropout} = 0.05$ for Llama-3.2.

We first create \ourDatasetShortName-induced variants of SeeClick \cite{cheng-acl24} and Llama3.2 \cite{meta-llama32}, named SeeClick-\ourDatasetShortName and Llama3.2-\ourDatasetShortName, by fine-tuning them on \ourDatasetShortName.
For SeeClick-\ourDatasetShortName, we fine-tune SeeClick for 10 epochs using the AdamW optimizer (learning rate: 1e-5, cosine decay, batch size: 16). The checkpoint from epoch 7 is selected.
For Llama3.2-\ourDatasetShortName, we fine-tune Llama3.2 for 10 epochs using AdamW (learning rate: 1e-4, StepLR with gamma: 0.85, batch size: 24). The checkpoint from epoch 10 is selected.

Next, both the original and \ourDatasetShortName-induced models are trained for 10 epochs on either of AitW and AMEX datasets using the AdamW (learning rate: 3e-5, batch size: 16). We select the checkpoint with the lowest validation loss for evaluation.
Learning rate schedulers follow the settings in their public repositories: cosine decay for SeeClick and StepLR (gamma: 0.85) for Llama-3.2.
Each training sample consists of:
\begin{compactitem}
\item Current screen image
\item Task description
\item Previous 4 actions as context (list of action types, coordinates, and typed texts)
\end{compactitem}

\subsection{Unifying the action space for comparison}

To evaluate the finetuned models on the AitW, AMEX, \ourDatasetShortName, and Windows Mobile test sets simultaneously, we focus on touch operations along with long press and typing actions.
These actions have clear one-to-one mappings between the datasets and represent fundamental mobile OS interactions, covering 78.51\% of the AitW test set, 82.60\% of the \ourDatasetShortName test set and 94.39\% of the Windows Mobile test set.
For touch actions, we evaluate coordinate predictions against ground truth interaction regions.
Typing actions are validated using flexible text matching, considering a prediction correct if the predicted text exactly matches the reference text or if either contains the other.
We believe this focused evaluation approach allows for meaningful comparisons while acknowledging the diverse interaction patterns across mobile platforms.

\subsection{Expanded Results}
On the AitW dataset, Table \ref{mobvid:supp_tab:cross_platform} expands on the summary results in Table \ref{mobvid:tab:cross_platform} by providing task-specific performance across five categories: General, Google (short for GoogleApps), Install, Shopping (short for WebShopping), and Single.
The results show that the \ourDatasetShortName-finetuned models consistently outperform the AitW-finetuned baselines in all evaluation categories, demonstrating their robustness in handling diverse tasks.

\begin{table*}[ht!]
\centering
\setlength\tabcolsep{3.8pt}
\begin{tabular}{lccccccccc}
\toprule
& \multicolumn{9}{c}{Test set} \\ \cline{2-10} \noalign{\vskip 1mm}
\multirow{1}{*}{Finetuned Models} & \multicolumn{6}{c}{AitW} & \multirow{2}{*}{AMEX} & \multirow{2}{*} {\ourDatasetShortName} & \multirow{2}{*}{Windows Mobile} \\
\cline{2-7}
 & General & Google & Install & Shopping & Single & Avg & & & \\
\midrule
\multicolumn{9}{l}{\textbf{\textit{AitW-finetuned from}}:} \\
\quad SeeClick &	\textbf{63.19}	&	67.21	&	\textbf{64.26}	&	78.98		&	61.25		&	66.98		&	47.23		&	40.66		&	38.54  \\
\quad SeeClick-\ourDatasetShortName &	62.83	&	67.21	&	64.09	&	\textbf{79.79}	&	\textbf{68.44}	&	\textbf{68.47}	&	\textbf{47.76}	&	\textbf{63.39}	&	\textbf{51.71} \\
\specialrule{0.1pt}{0.5pt}{1.5pt}
\multicolumn{9}{l}{\textbf{\textit{AMEX-finetuned from}}:} \\
\quad SeeClick &	33.45		&	\textbf{46.72}		&	32.41		&	33.45		&	39.38		&	37.08		&	\textbf{68.19}		&	44.23		&	43.17 \\
\quad SeeClick-\ourDatasetShortName &	\textbf{35.04}	&	40.98	&	\textbf{35.42}	&	\textbf{42.62}	&	\textbf{46.88}	&	\textbf{40.19}	&	66.13	&	\textbf{63.39}	&	\textbf{55.37} \\
\midrule
\multicolumn{9}{l}{\textbf{\textit{AitW-finetuned from}}:} \\
\quad Llama-3.2 &	55.93		&	63.45		&	58.08		&	68.87		&	48.44		&	58.96		&	43.74		&	39.80		&	26.83 \\
\quad Llama-3.2-\ourDatasetShortName &	\textbf{61.95}	&	\textbf{70.68}	&	\textbf{67.18}	&	\textbf{76.77}	&	\textbf{60.31}	&	\textbf{67.38}	&	\textbf{55.96}	&	\textbf{57.99}	&	\textbf{50.24} \\
\specialrule{0.1pt}{0.5pt}{1.5pt}
\multicolumn{9}{l}{\textbf{\textit{AMEX-finetuned from}}:} \\
\quad Llama-3.2 &	28.67		&	29.32		&	27.54		&	31.94		&	31.56		&	29.81		&	61.30		&	40.17		&	28.29 \\
\quad Llama-3.2-\ourDatasetShortName &	\textbf{37.88}	&	\textbf{42.57}	&	\textbf{38.83}	&	\textbf{49.59}	&	\textbf{45.94}	&	\textbf{42.96}	&	\textbf{72.36}	&	\textbf{58.35}	&	\textbf{51.46} \\
\bottomrule
\end{tabular}
\caption{Comparison of navigation action accuracies with the original pre-trained models (SeeClick, Llama-3.2) vs. the corresponding \ourDatasetShortName-induced variants (SeeClick-\ourDatasetShortName, Llama-3.2-\ourDatasetShortName).
Results on AitW \cite{rawles-neurips23} test set are broken down by their original evaluation categories alongside overall averages.
The \ourDatasetShortName-induced variants mostly achieve higher performance across different mobile platforms, including significantly better adaptation to the previously unseen mobile platform (Windows Mobile).
}
\label{mobvid:supp_tab:cross_platform}
\end{table*}

%% file: review/supp/expanded_related.tex
\subsection{Cross-Domain GUI Datasets and Approaches}

Early GUI agent benchmarks often focused on simple, single-step tasks or were confined to a single platform, limiting cross-environment generalization \cite{shi-pmlr17}. 
Recently, there have been efforts to scale up data collection for web and GUI-based environments to support the training of agents on a wider range of computer interaction tasks.
AgentTrek \cite{xu-arxiv24} simulates actions in a virtual environment based on tutorial text, with step-by-step instructions from GPT-4o. WebArena \cite{zhou-iclr23} offers a high-fidelity browser simulation with complex, long-horizon web tasks. Mind2Web \cite{deng-neurips23} collects crowdsourced demonstrations on real-world websites, though data collection is expensive and limited to web domains. GUI-World \cite{chen-iclr2025} spans multiple platforms (web, mobile, desktop), but is restricted to question-answering rather than full action-based tasks.

While simulator-based approaches in web and computer OS domains can extend to Android via emulators, iOS's closed APIs hinder automated interaction extraction, limiting multi-platform coverage.
\ourDatasetShortName avoids direct GUI access by leveraging YouTube videos and automatically detecting scenes and actions.
Unlike simulators, which provide built-in interaction logs, \ourDatasetShortName tackles data extraction using OCR-based scene segmentation, UI detection via GroundingDINO, and a three-step action identification pipeline.
This approach is also adaptable to web and desktop GUIs, although higher resolutions and complex interactions may introduce new challenges.

%% file: review/supp/episode_examples.tex
\begin{figure*}[h]
\centering
\includegraphics[width=0.85\linewidth]{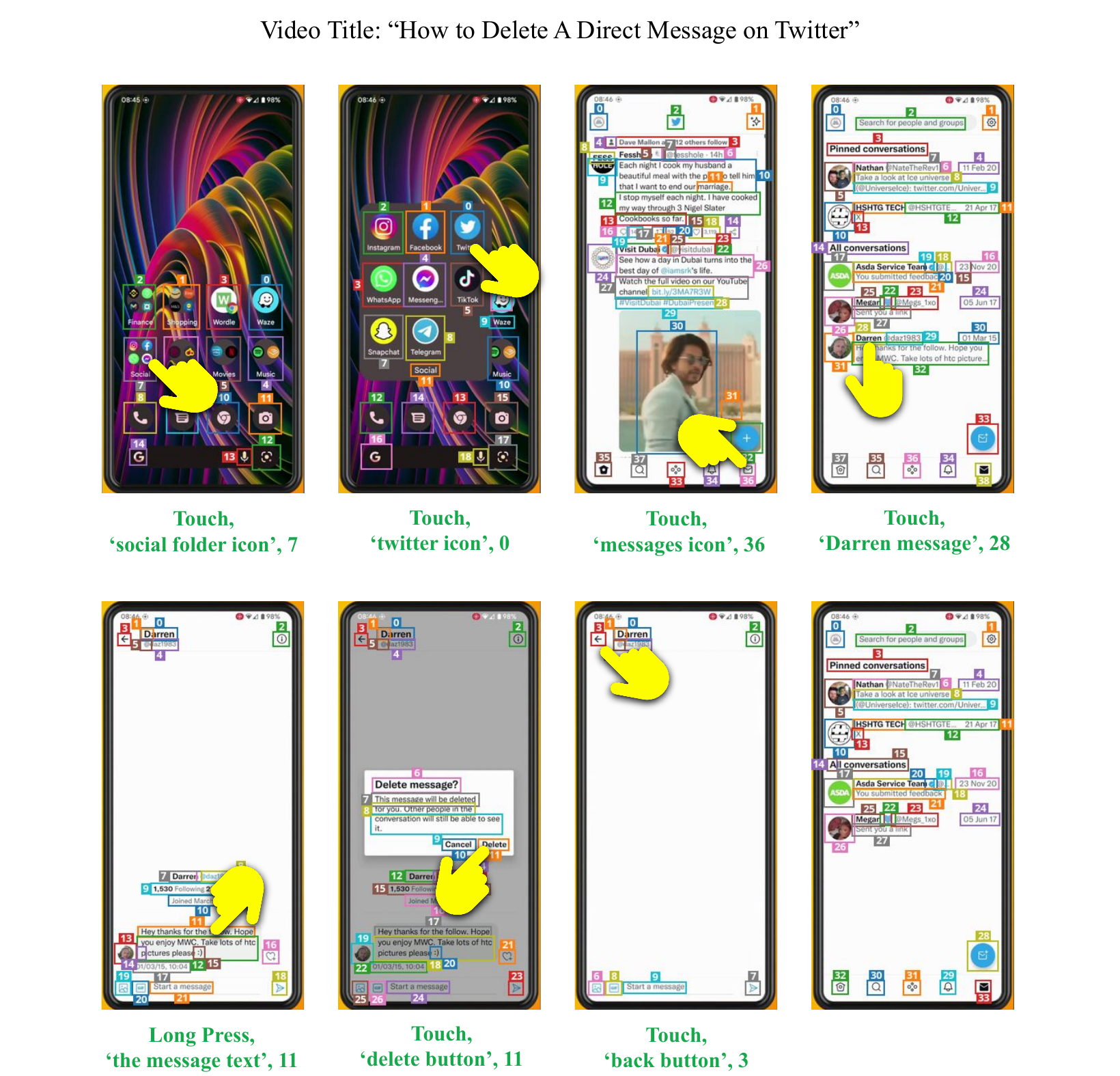}
\caption{Example of perfect action identification for deleting a direct message on Twitter in Android. Each touch and long press action is annotated with the corresponding box ID and visual indicator.}
\label{mobvid:suppfig:episode_allcorrect_android}
\end{figure*}
\begin{figure*}[h]
\centering
\includegraphics[width=0.85\linewidth]{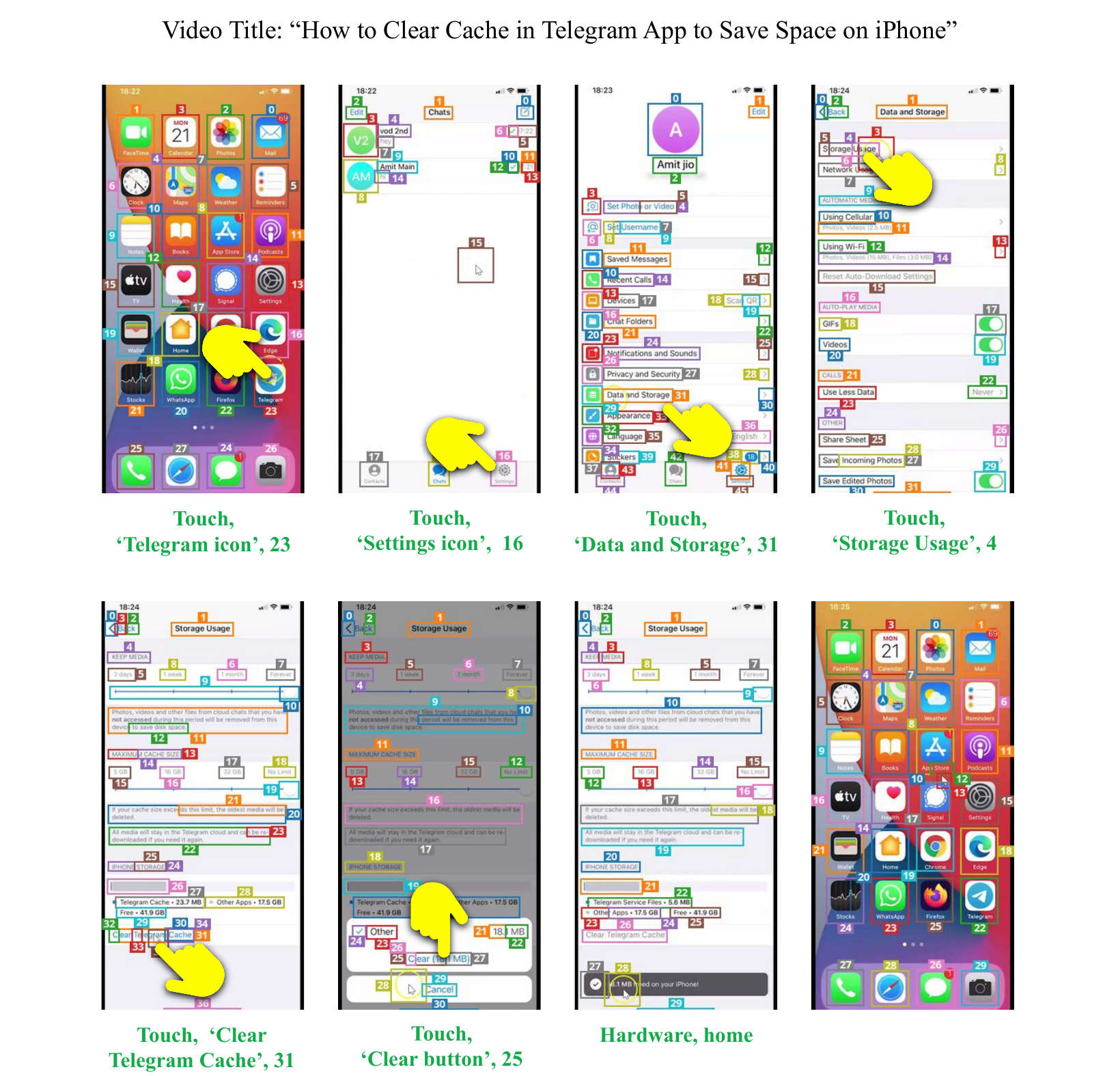}
\caption{Example of perfect action identification for clearing Telegram cache on iOS. Each touch action is labeled with the corresponding box ID and highlighted with a visual indicator.}
\label{mobvid:suppfig:episode_allcorrect_ios}
\end{figure*}
\begin{figure*}[h]
\centering
\includegraphics[width=0.85\linewidth]{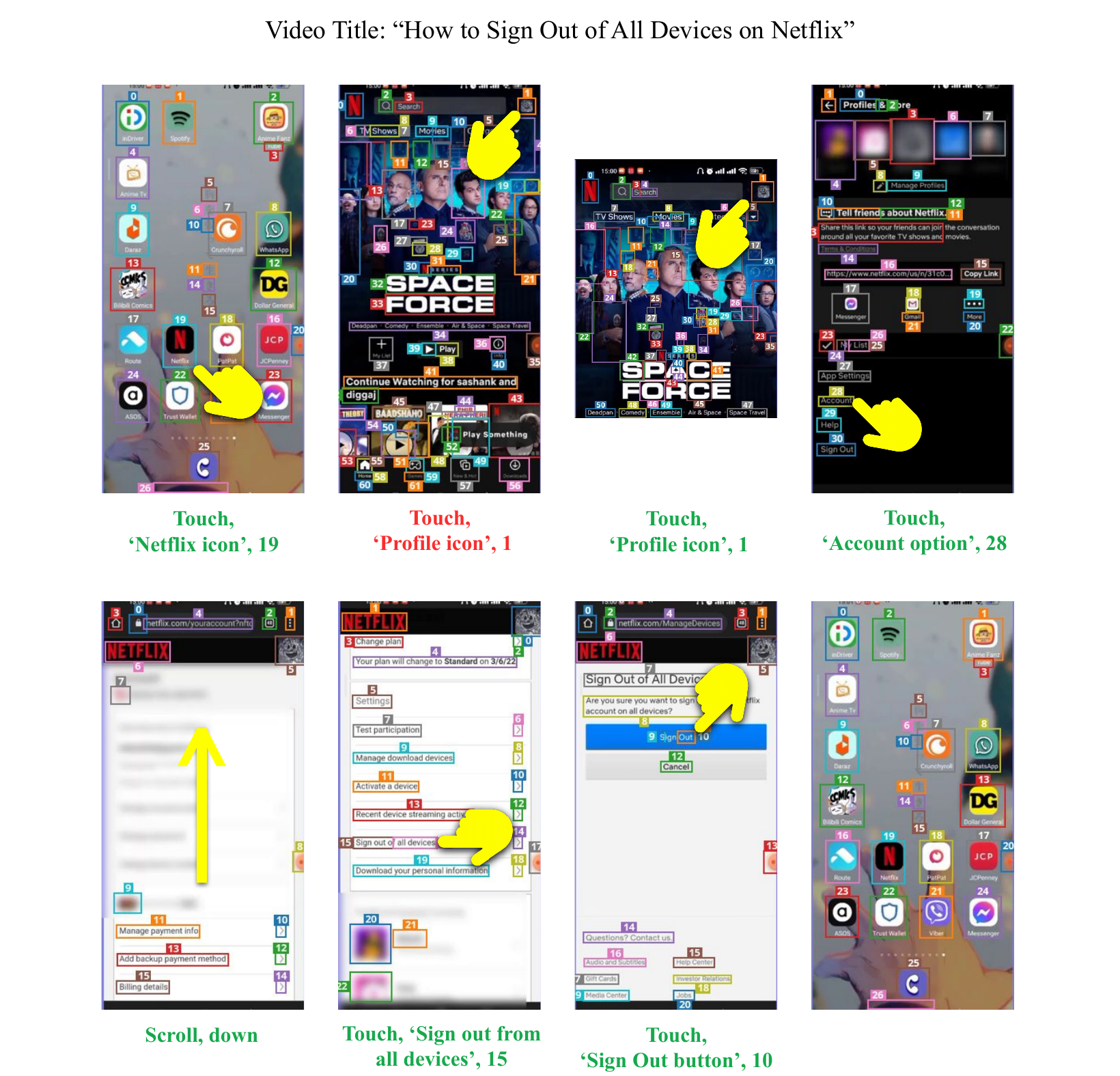}
\caption{Example showing path selection behavior for signing out of all Netflix devices on Android. \textcolor{customgreen}{Green} indicates correct actions, \textcolor{customred}{red} indicates alternate valid actions that could achieve the same goal.}
\label{mobvid:suppfig:episode_nearmiss1_android}
\end{figure*}
\begin{figure*}[h]
\centering
\includegraphics[width=0.85\linewidth]{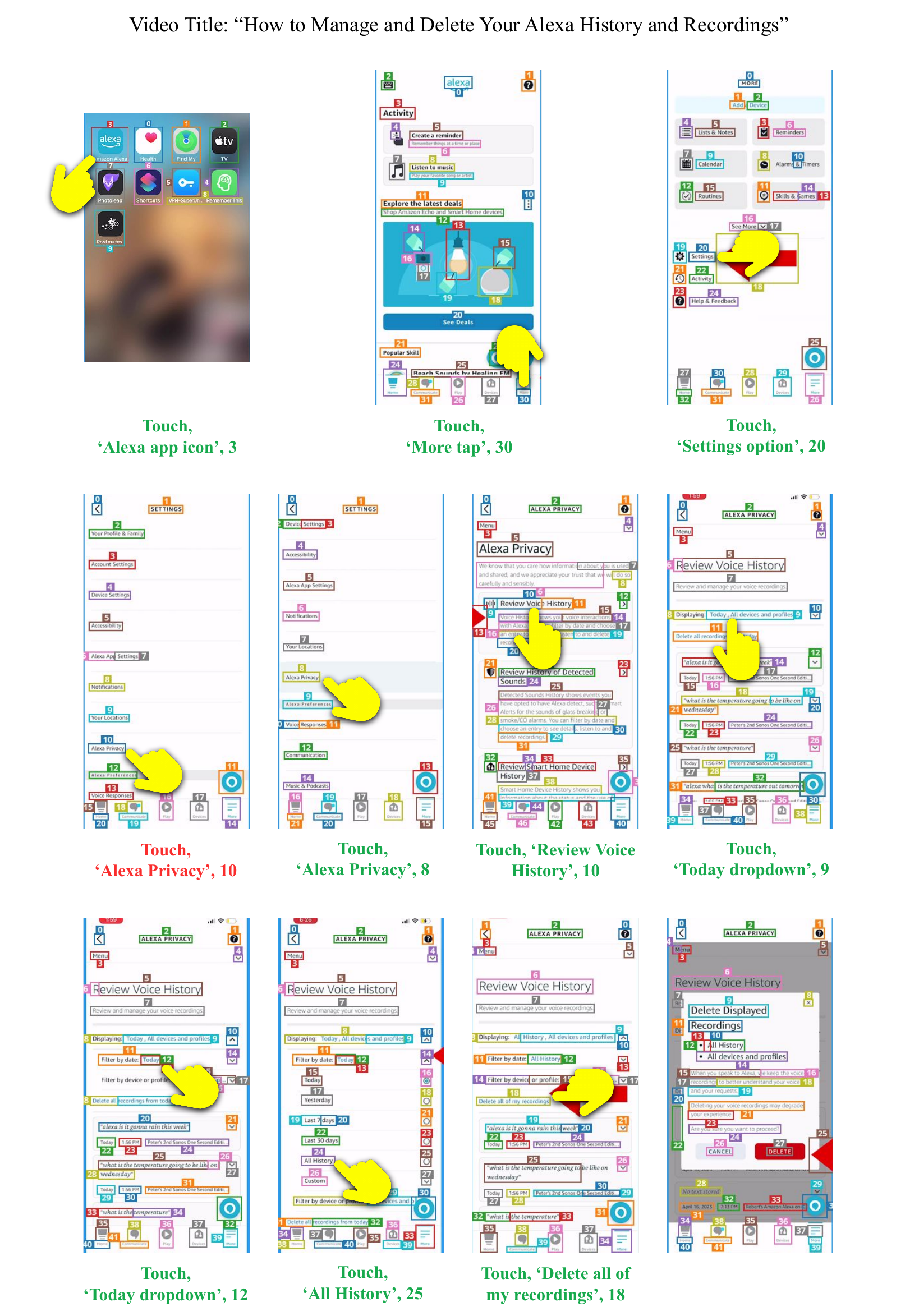}
\caption{Example showing path selection behavior for managing Alexa history and recordings on iOS. \textcolor{customgreen}{Green} indicates correct actions, \textcolor{customred}{red} indicates alternate valid approaches that were not selected.}
\label{mobvid:suppfig:episode_nearmiss1_ios}
\end{figure*}
\begin{figure*}[h]
\centering
\includegraphics[width=0.85\linewidth]{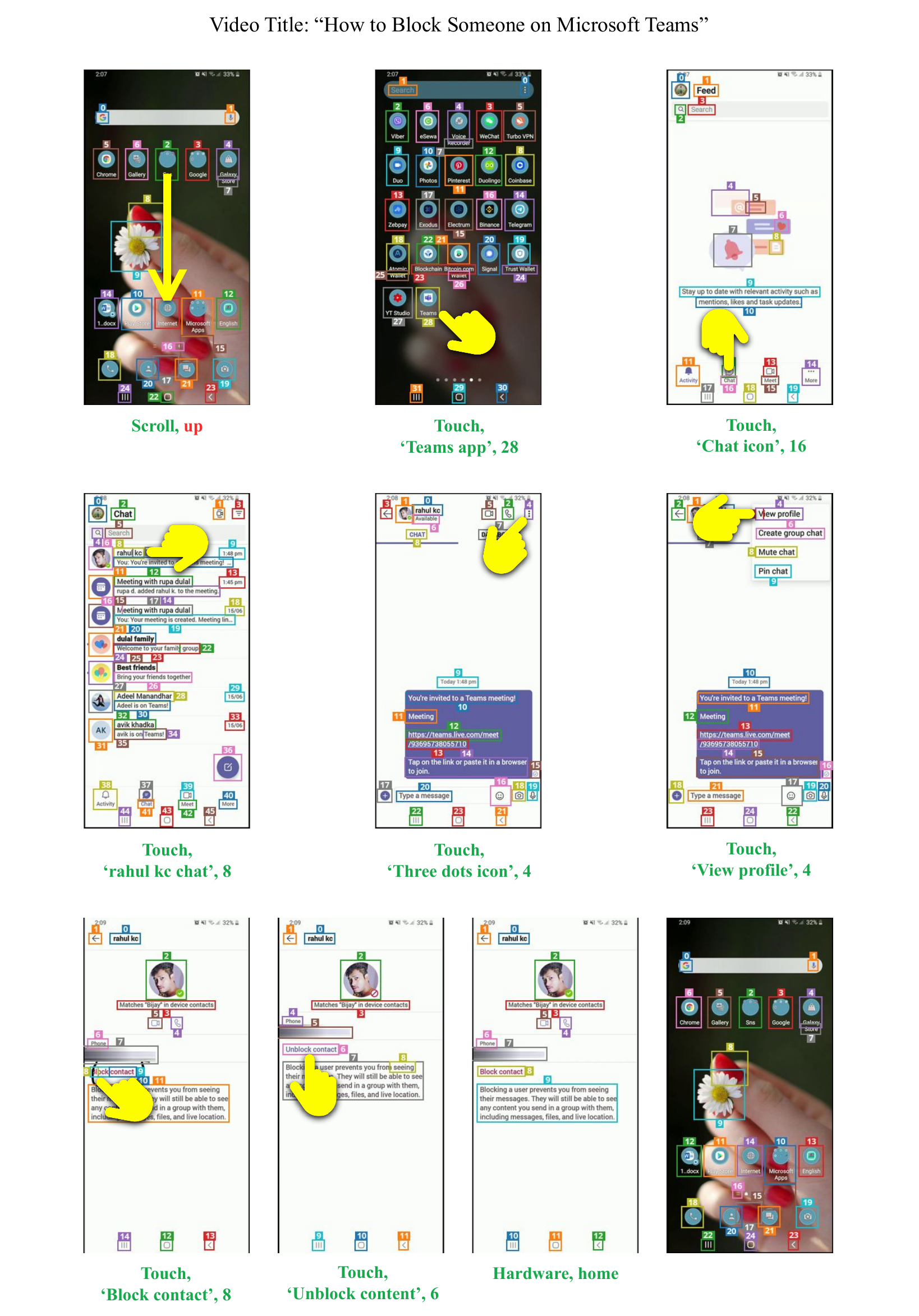}
\caption{Example identifying scrolling direction in Android. \textcolor{customgreen}{Green} indicates correct actions, \textcolor{customred}{red} shows incorrect scrolling direction prediction.}
\label{mobvid:suppfig:episode_nearmiss2_android}
\end{figure*}
\begin{figure*}[h]
\centering
\includegraphics[width=0.85\linewidth]{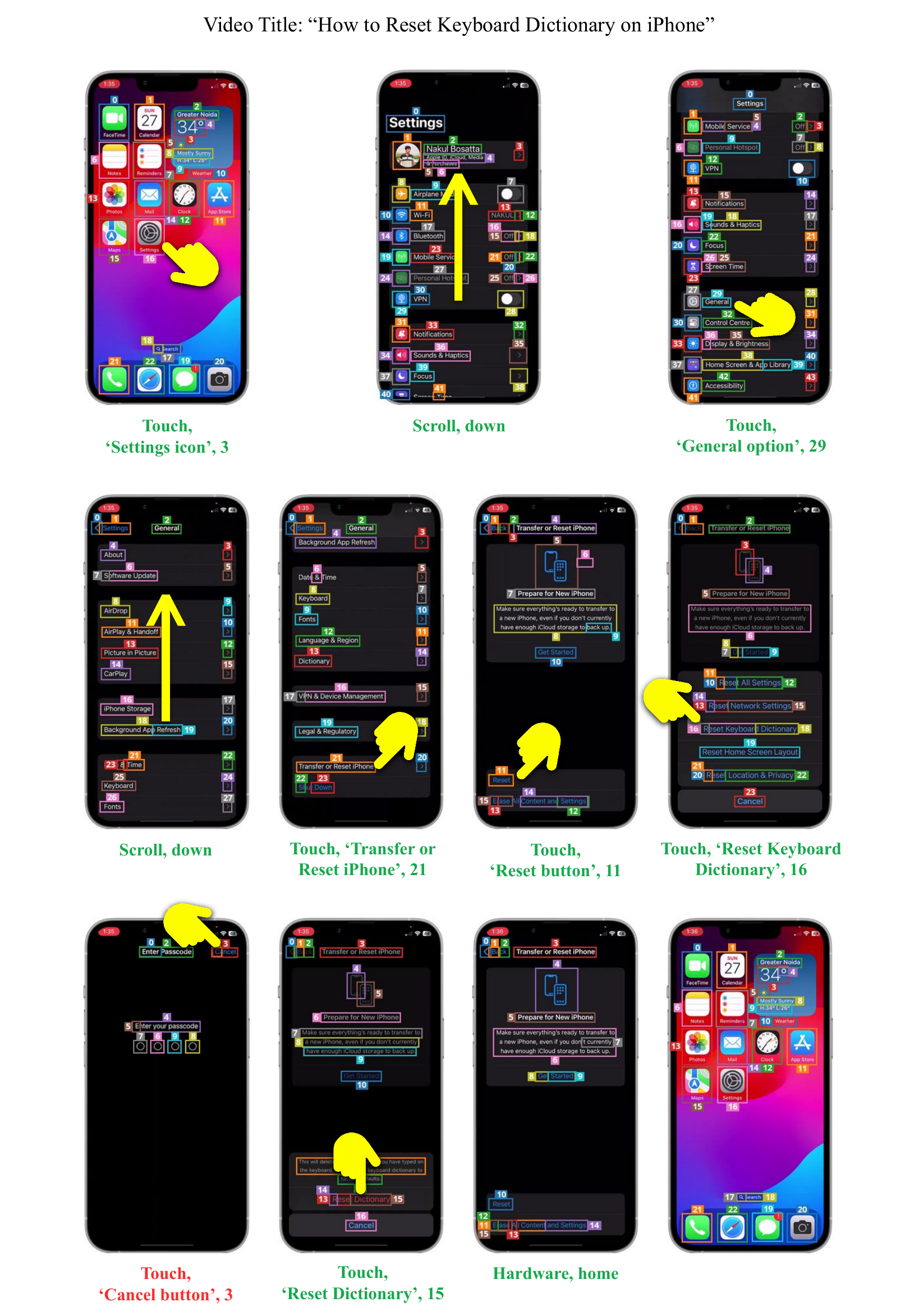}
\caption{Example showing handling of authentication challenges when resetting keyboard dictionary on iOS. \textcolor{customgreen}{Green} indicates correct actions, \textcolor{customred}{red} shows where the system selected cancel instead of handling passcode entry.}
\label{mobvid:suppfig:episode_nearmiss2_ios}
\end{figure*}

We present example episodes from our dataset to demonstrate the effectiveness of our action identification framework. 
The examples are organized into three categories: perfectly identified sequences, near-miss cases with multiple valid action paths, and challenging cases involving platform-specific operations.
Figures \ref{mobvid:suppfig:episode_allcorrect_android} and \ref{mobvid:suppfig:episode_allcorrect_ios} showcase successful action identification sequences on Android and iOS platforms, respectively. 
In these examples, our framework correctly identifies all user interactions, demonstrating its robustness across different mobile operating systems.
Figures \ref{mobvid:suppfig:episode_nearmiss1_android} and \ref{mobvid:suppfig:episode_nearmiss1_ios} illustrate cases where multiple valid interaction paths exist. 
Our framework typically selects the most direct path to accomplish the task, though this may occasionally differ from human demonstrations.
Figures \ref{mobvid:suppfig:episode_nearmiss2_android} and \ref{mobvid:suppfig:episode_nearmiss2_ios} present challenging scenarios involving platform-specific operations or security features. 
These examples highlight current limitations in handling specialized interactions like secure input or complex scrolling patterns.

